\documentclass{article}

\PassOptionsToPackage{inline}{enumitem}
\PassOptionsToPackage{most}{tcolorbox}
\usepackage[arxiv1col,nonatbib,notheorems,colorlinks]{styles/paperkit}

\usepackage[
  backend=biber,
  datamodel=acmdatamodel,
  style=acmnumeric,
  natbib=true,
  ]{biblatex}
\usepackage{common}

\let\dcwrapfigure\wrapfigure
\let\dcwraptable\wraptable
\renewcommand{\wrapfigure}[1]{\dcwrapfigure{R}}
\renewcommand{\wraptable}[1]{\dcwraptable{R}}
\usepackage{needspace}

\tikzset{every picture/.append style={execute at begin picture={%
  \renewcommand{\scriptsize}{\fontsize{7}{8}\selectfont}%
  \renewcommand{\footnotesize}{\fontsize{8}{9.5}\selectfont}}}}

\providecommand{\Description}[2][]{}

\newcommand{\aratest}{\textsc{ARATest}\xspace}
\newcommand{\dissclaimer}{\textsc{DissClaimer}\xspace}

\papertitle{%
  From Prose to Properties:
  Testing Research Claims with \aratest for Agent-Native Science%
}
\paperrunningtitle{Testing Research Claims with ARATest for Agent-Native Science}

\paperauthors{%
  Yifeng He\textsuperscript{1,\,2},\ %
  Jiachen Liu\textsuperscript{2},\ %
  Hao Chen\textsuperscript{3}%
}
\paperaffiliations{%
  \textsuperscript{1}University of California, Davis,\ %
  \textsuperscript{2}ARA Lab,\ %
  \textsuperscript{3}The University of Hong Kong%
}
\paperabstract{AI agents now run experiments and report findings, so research
produces claims faster than researchers can check them. Each
empirical claim implies a property that must hold over its cases, but it
states that property only in prose. Without an explicit property, no test can
decide which evidence would falsify a claim, so validation cannot keep pace
with the claims agents produce.
To make claims testable, we approximate a claim's semantics by abstracting its
\emph{property skeleton}, a parameterized relation that recurs across claims.
We present \aratest, a property-based testing library that checks
these skeletons as executable tests.
A typed declaration binds a claim to a skeleton and fixes the claim's
compared quantities, cases, scope, and numerical policy. A shared case type
makes every generator supply cases in the claim's statistical unit.
A three-valued semantics assigns each property one of three verdicts: holds,
violated, or inconclusive. Every violation carries a replayable witness.
A default generator reads the evidence that agent-native research artifacts
(\ara) link to each claim, and auto-research generators dispatch an agent to
run released code for fresh cases. Our \dissclaimer plugin supplies that
agent and runs these generators as continuous testing.

Of 243 claims in 33 existing \ara artifacts, 54 received a faithful
declaration bound to evidence. \aratest decided 40 of them: 32 held and
eight were violated, and the other 14 remained inconclusive.
Two violations exposed errors in published papers; we found no
prior report of either, and the authors of both papers confirmed them. In
each, the prose contradicts the paper's own table.
\aratest detected all \num{553} seeded
faults in reported tables without a false alarm on 19 valid controls, whereas
two LLM judges given the same evidence detected 342 and 319.
On fresh runs of released code from two peer-reviewed papers,
four of six tests held, and the other two exposed behaviors that the
reported evidence could not show.
In \dissclaimer campaigns on three of the 33 artifacts, six of 11 checks
held and none was violated.
These results suggest that once a claim's property is explicit, one fixed
test can check the claim against both reported and fresh evidence and catch
errors that LLM judges miss.
}
\papercorrespondence{%
  Yifeng He (\href{mailto:yfhe.cs@gmail.com}{yfhe.cs@gmail.com})
}
\papercode{\url{https://github.com/ARA-Labs/aratest}}

\paperlogo{}{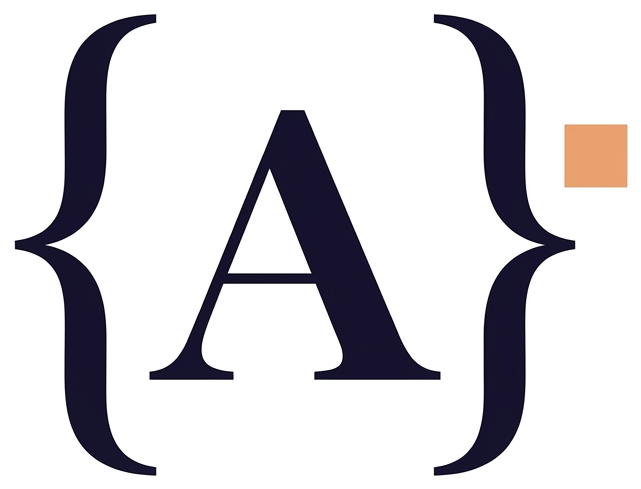}
\paperlogoheight{20pt}
\paperbrand[11pt]{figures/ara-logo.png}{ARA Lab}
\papershortauthors{Yifeng He, Jiachen Liu, and Hao Chen}

\begin{document}

\makepaperheader

\section{Introduction}
\label{sec:intro}

Over the past two decades, high-level languages and open-source libraries
made writing a program cheap, so the bottleneck shifted from producing
software to establishing that it is correct~\cite{kim2026keynote}. AI agents
now bring the same shift to research, the other half of the research and
development cycle.
They design experiments, run them, and report
results~\cite{ghareeb2026robin,gottweis2026coscientist,aygun2026era},
and even use those results to improve themselves~\cite{favaro2026rsi,duan2026lastai}.
\citet{openai2026acceleration} reported 3.1 agent-workdays of runtime for every
workday of human research labor, and venue submissions keep
growing~\cite{iclr2026retrospective}. Once producing a result stops being the
bottleneck, deciding which results justify the next step becomes one.
That decision requires validating each claim against its evidence. But
a claim states the property it implies only in prose, so automated validation
falls to LLM judges that read the prose~\cite{zheng2023judging}. This check
is unsound and scales poorly: LLM judges are slow and costly, and the
same evidence can pass on one run and fail on the
next~\cite{stureborg2024inconsistent,schroeder2025trust}.

Software analysis scales because it checks a program against an explicit
representation of what the program should do, kept apart from the program
itself. Abstract interpretation checks a program through an abstraction of
its semantics~\cite{cousot1977abstract}, and testing checks it on concrete
executions against a stated property. Property-based testing (PBT) separates
the two concerns: a property states what must hold, and a generator supplies
the cases~\cite{claessen2000quickcheck}. A research claim keeps its
property inside its prose, with no separate representation. To give claims
that representation, we approximate
the semantics of research claims by extracting their \emph{abstract
properties}, the relations a claim asserts independent of the evidence that
instantiates them. We give these properties typed, executable
specifications that a PBT framework can check.

\begin{wraptable}{r}{0.45\columnwidth}
  \centering
  \wraptop
  \caption{Claim C01's scope boundary in \emph{LCA-on-the-Line}~\cite{shi2024lca}.}
  \label{tab:worked}
  \wrapcapskip
  \renewcommand{\arraystretch}{1.15}
  \footnotesize
  \resizebox{\linewidth}{!}{\begin{tabular}{@{}l l c c c@{}}
    \toprule
    & Dataset & \shortstack{LCA\\ \scriptsize subject} & \shortstack{Top1\\ \scriptsize reference}
      & \shortstack{LCA $>$ Top1\\ \scriptsize (\texttt{gt})} \\
    \midrule
    \multirow{4}{*}{\rotatebox[origin=c]{90}{\footnotesize in scope}}
      & ImgN-S  & 0.903 & 0.275 & \holds \\
      & ImgN-R  & 0.883 & 0.140 & \holds \\
      & ImgN-A  & 0.839 & 0.094 & \holds \\
      & ObjNet  & 0.956 & 0.522 & \holds \\
    \midrule
    \footnotesize widen
      & ImgN-v2 & 0.582 & 0.980 & \flips{-0.398} \\
    \bottomrule
\end{tabular}
}
  \wrapbottom
\end{wraptable}
An abstract property is useful only if it says exactly what the claim says.
In \emph{LCA-on-the-Line}~\cite{shi2024lca}, one claim states that
in-distribution LCA distance predicts out-of-distribution accuracy more
strongly than Top-1 accuracy on four named shifted datasets. The reported
correlations support the comparison on all four but reverse on ImageNet-v2,
which the claim explicitly excludes (\autoref{tab:worked}). A property that
quantified over every dataset in the table would call the claim false, while
a property that ignored the dataset list could not tell a boundary from a
counterexample. A reusable test must therefore pin down both the comparison
and the scope of observations that can decide its verdict. Without such a
test, each new assessment of the claim can answer a different question.

We present \aratest, a PBT library that turns the abstract properties
of research claims into reusable executable tests. \aratest provides
abstract properties as \emph{property skeletons}, parameterized relations that
recur across claims. Skeletons build on property-specification
patterns~\cite{dwyer1999patterns} and on recent property templates for
testing and proof~\cite{qiu2026tensor,lee2026templates}. For each claim, a declaration binds a skeleton
to a contract that fixes the compared quantities, quantified cases,
admissible scope, and numerical policy (\autoref{sec:contract}).
In a typed declaration, \Kind selects a skeleton and \Domain fills that
skeleton's parameters. In our design, the research agent writes this declaration
while it does the research and records it with the claim in its agent-native
research artifact (\ara)~\cite{liu2026ara}.

\aratest keeps the property fixed and lets generators supply cases. 
Its default generator reads reported evidence from the \ara,
which links claims to their evidence. Its auto-research generators draw fresh cases from
experiment families that the research agent binds to a claim when it
states the claim. For each such generator, \aratest dispatches an experiment-generator agent that runs the family's experiments
on released code. \aratest defines only the interface between generator and
agent, so any agent that returns cases of the property's type can serve.
Our \dissclaimer plugin supplies one such agent, built on
OpenHands~\cite{wang2025openhands}, and a service that runs these generators
as continuous testing campaigns under a budget.
Because the agent's run may not test the claim faithfully,
\aratest treats a failing case on this path as feedback, not a refutation (\autoref{sec:execution-backed}). 
Each property and its generators share a case type, such as a pair of table cells or a batch of
seeded runs. Every generator therefore supplies cases in the claim's
statistical unit. We define a three-valued verdict semantics (\autoref{sec:soundness}). 
We show that a violation's witness alone suffices
to replay the violation, while a pass corroborates only the cases checked.
\aratest reports a violation as a warning, and its user decides whether
the violation reveals a defect.

We evaluate \aratest on the compiled \ara artifacts of the
\ara-PaperBench corpus~\cite{liu2026ara,paperbench2025}, which includes our
\mmdfuse~\cite{biggs2023mmdfuse} artifact, and on one more artifact for
\simformer~\cite{gloeckler2024allinone}. We compiled the \mmdfuse and \simformer artifacts with the
\CompilerSkill skill from each paper and its released code
(\autoref{sec:evaluation}). Of the 243 claims in 33 existing
artifacts, 55 received a declaration that passed fidelity review, and 54 fully
bound to reported evidence (22.2\%). \aratest decided 40 claims (16.5\%): 32 held and
eight contradicted their mapped evidence. Excluding five decisions that
check only reported summaries leaves 35 (14.4\%).
A source triage traced two violations to errors in published papers; we
found no prior report of either, and the authors of both papers confirmed
them. In each, a sentence states numbers that the
paper's own table contradicts. Four violations instead came from the
compiled artifact. Fourteen bound claims
remain inconclusive. The other 189 claims lack complete binding although
all 33 artifacts load, and most of them need a verification method other
than property testing. Because we designed the catalog on these claims,
these results bound what the current catalog reaches; they do not measure
scientific validity or extraction quality on unseen papers
(\autoref{sec:eval-fidelity}).

\aratest detected all \num{553} faults
that we seeded into reported tables and raised no false alarm on 19 valid
controls. On the same evidence, a direct LLM judge detected 342 faults and a
calculator-assisted judge 319, and they took nine to ten times longer per
check (\autoref{sec:eval-detection}).
Replay of an unseeded finding traced an inconsistency in the compiled \sema
table to the \CompilerSkill skill swapping two metric headers, while the published
claim held (\autoref{sec:eval-repeat}).
We also ran six property-based tests on the released code of
two peer-reviewed papers. Their generators instantiate four family types, and
the experiment-generator agent ran their experiments.
Four of the six tests held on fresh runs. The other two exposed
behaviors that the reported evidence could not show, but their registered
rules left both tests \vinconclusive (\autoref{sec:eval-execution}).
In \dissclaimer campaigns on three of the 33 artifacts, six of 11 checks
held and none was violated.
Our contributions are:
\begin{itemize}
\item An approach that approximates the semantics of research claims with
  abstract properties, fixed by a typed contract over relation, cases, scope,
  and numerical policy, and decided by a three-valued verdict semantics with
  replayable violation witnesses (\autoref{sec:contract} and
  \autoref{sec:soundness}).
\item \aratest, a testing library that binds reusable property
  skeletons to \ara claims, checks them against
  experimental evidence, and returns replayable witnesses
  (\autoref{sec:approach} and \autoref{sec:usage}).
\item Auto-research generators built from typed experiment families, each
  with its own scope check, that dispatch an experiment-generator agent to
  supply fresh experimental cases for the same property, and \dissclaimer,
  a plugin that supplies this agent and runs the generators as continuous
  testing (\autoref{sec:execution-backed}).
\item An evaluation of declaration fidelity on existing artifacts,
  seeded-fault detection, and experiments that the
  experiment-generator agent ran on released code
  (\autoref{sec:evaluation}).
\end{itemize}

\section{Background}
\label{sec:background}

\subsection{Structured research artifacts}
\label{sec:background:ara}

Agent-native research artifacts (\ara)~\cite{liu2026ara} provide a typed host
for the claims and evidence that \aratest tests. They link claims and
experiment plans in \texttt{logic/} to code, configurations, and environments
in \texttt{src/}, results and logs in \texttt{evidence/}, and research
history in \texttt{trace/}. The \CompilerSkill skill builds these records
from an existing paper or repository, while \LiveResearchManager records
decisions, experiments, and abandoned directions during an agent's research
loop. Provenance tags distinguish human input, model proposals, and executed
work. \RigorReviewer assesses the resulting artifact. The \ara format and
recording process are existing infrastructure. \aratest runs as another
skill that adds executable claim checks to this workflow.

Each claim entry in \texttt{logic/claims.md} records the assertion
(\Statement), its scope (\Conditions), what would disprove it
(\FalsCriteria), and its current standing (\Status). \Proof links to
supporting experiments, \EvidenceBasis describes their evidence, and
\Dependencies names related claims. These fields help a reader or agent find
the supporting work. They do not yet specify an executable oracle, a check that decides whether
evidence passes.
\aratest adds a declaration that states this oracle as a typed
contract. The declaration also retains the source claim and its provenance,
so a reviewer can check whether the binding is faithful.

\subsection{Property-based testing and property skeletons}

Property-based testing (PBT) checks a program against a stated property on
many generated inputs. \citet{claessen2000quickcheck} introduced PBT with
QuickCheck, a Haskell library that tests functions against properties written
as ordinary Haskell functions. A property-based test has three parts. A
\emph{property} is a Boolean function that should return true on every input.
For example, reversing a list twice should restore the list. A \emph{generator}
supplies inputs. QuickCheck derives a default generator from the property's
argument type, here random lists of integers, and users can write custom
generators for the data a test requires. The \emph{oracle} decides whether an
input passes. In QuickCheck, the property serves as its own oracle, so a test
fails when the property returns false. The failing input is a concrete
counterexample that a developer can inspect and replay. If a developer
mistakenly states that a single reversal leaves every list unchanged, QuickCheck
reports a failing list such as \texttt{[0,1]}. When every generated case
passes, the result is evidence for those cases only. It does not prove the
property for all inputs.

Many properties share one shape and differ only in what they apply to. A
\emph{property skeleton} states that shape once as a parameterized property,
and instantiating the skeleton fills its parameters to yield a concrete test.
\citet{qiu2026tensor} use skeletons to test tensor compilers and specify each
skeleton by a relation, its applicability conditions, a way to construct test
instances, and a rule for comparing outputs. A commutativity skeleton, for
example, compares $f(x,y)$ with $f(y,x)$. To instantiate it, a tester selects
an operator for which commutativity holds, such as addition but not
subtraction, generates compatible operands, and chooses exact equality or a
floating-point tolerance. The relation alone therefore does not define a
valid test. The applicability conditions and the comparison rule belong to
the skeleton as much as the relation does.

Research claims also recur in a few shapes. A comparative claim, for example,
asserts that one quantity exceeds another on a set of cases. Unlike
a commutativity skeleton, such a claim also has a \emph{scope}, the cases it covers. A
reversal outside that scope marks the claim's boundary rather than a
counterexample.

\section{Property-based testing for research claims}
\label{sec:approach}

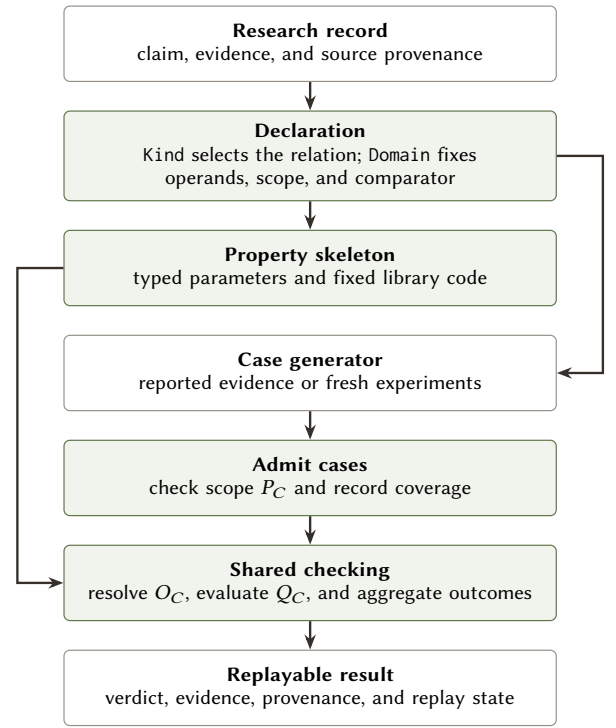
\begin{wrapfigure}{r}{0.45\textwidth}
  \vspace{-3em}
  \centering
  \resizebox{\linewidth}{!}{\begin{tikzpicture}[
  font=\sffamily\scriptsize,
  box/.style={draw=oasiswarmgray, rounded corners=2pt, align=center,
    inner sep=4pt, minimum height=0.9cm, text width=5.6cm, fill=white},
  fixed/.style={box, draw=oasisolive, fill=oasisolivebg},
  flow/.style={-{Stealth[length=1.8mm]}, thick, draw=oasischarcoal}
]
  \node[box] (source)
    {\textbf{Research record}\\claim, evidence, and source provenance};
  \node[fixed, below=0.34cm of source] (declaration)
    {\textbf{Declaration}\\\Kind selects the relation; \Domain fixes operands, scope, and comparator};
  \node[fixed, below=0.34cm of declaration] (skeleton)
    {\textbf{Property skeleton}\\typed parameters and fixed library code};
  \node[box, below=0.34cm of skeleton] (generator)
    {\textbf{Case generator}\\reported evidence or fresh experiments};
  \node[fixed, below=0.34cm of generator] (admission)
    {\textbf{Admit cases}\\check scope $P_C$ and record coverage};
  \node[fixed, below=0.34cm of admission] (checker)
    {\textbf{Shared checking}\\resolve $O_C$, evaluate $Q_C$, and aggregate outcomes};
  \node[box, below=0.34cm of checker] (result)
    {\textbf{Replayable result}\\verdict, evidence, provenance, and replay state};

  \draw[flow] (source) -- (declaration);
  \draw[flow] (declaration) -- (skeleton);
  \draw[flow] (declaration.east) -- ++(0.55,0) |- (generator.east);
  \draw[flow] (skeleton.west) -- ++(-0.55,0) |- (checker.west);
  \draw[flow] (generator) -- (admission);
  \draw[flow] (admission) -- (checker);
  \draw[flow] (checker) -- (result);
\end{tikzpicture}}
  \caption{System overview. The declaration fixes a claim-specific property.
    Either generator supplies cases; checking evaluates them and
    retains the deciding evidence.}
  \Description{A research record supplies a typed declaration, which fixes the
    property. A generator supplies reported or freshly measured cases.
    Checking admits cases, evaluates the property, and records a verdict
    with evidence for replay.}
  \label{fig:pipeline-overview}
  \wrapbottom
\end{wrapfigure}

\aratest makes a research claim reusable as a test. A declaration fixes
the property that the claim implies, a property skeleton checks it, and a generator supplies
cases. One generator reads reported evidence, and auto-research generators
produce fresh measurements from experiments. Every generator feeds the same
property without changing its operands, scope, or decision rule.
For claim C01 (\autoref{tab:worked}), the declaration limits the
property to LCA and Top-1 correlations on four named datasets, so
\aratest does not treat every number in the artifact as evidence about
the claim.

We borrow this split between a fixed property and varying cases from
PBT, where a user writes a property once and generators vary its
inputs~\cite{claessen2000quickcheck}. The
declaration approximates the claim's semantics with that property: it retains the
metric, method variant, baseline cohort, and experimental conditions needed to
interpret a comparison. \aratest fixes these commitments before any
generator supplies observations (\autoref{fig:pipeline-overview}).

The two kinds of generator support two modes of testing an artifact.
\emph{Integration testing} runs the reported-evidence generator when a
researcher or a research agent integrates new results into an \ara
artifact. Because it reads only stored evidence, it returns a verdict while
the results are being integrated, much as continuous integration tests a code
change before it merges. The \aratest library runs this mode through its
\texttt{aratest test} command (\autoref{sec:usage}). \emph{Continuous testing} runs auto-research
generators asynchronously and offline, much as OSS-Fuzz fuzzes open-source
projects continuously~\cite{ossfuzz2016}. Its experiments can take hours, so
its failing cases return to the research agent later as feedback instead of
blocking an integration. Our \dissclaimer plugin runs this mode. It supplies
an experiment-generator agent and a service that schedules the agent's runs
under a budget. \dissclaimer saves every violation as a record that
\aratest replays without the plugin.

We define the claim-testing contract (\autoref{sec:contract}), abstract
reusable property skeletons (\autoref{sec:router}), and show how a declaration
instantiates a skeleton from an \ara claim (\autoref{sec:schema}). We then
present the reported-evidence generator (\autoref{sec:catalog}) and the
auto-research generators (\autoref{sec:execution-backed}), and discuss how to
interpret a verdict that \aratest produces (\autoref{sec:soundness}).

\subsection{The claim-testing contract}
\label{sec:contract}

A claim $C$ becomes a test through three commitments. A domain predicate
$P_C$ defines which cases may test it. An observation map $O_C$ selects the
quantities that the claim concerns. A relation $Q_C$ states the required
comparison, including its quantifier, aggregation, and numeric or uncertainty
policy. A case $x$ takes the claim's statistical unit: it can
be one pair of table cells, a matched pair of runs, or an entire seed family
when the claim concerns an aggregate.
Keeping that unit explicit prevents a generator from replacing a claim about
a mean with a per-seed requirement.

Let $\mathcal{R}_A^g(x)$ denote the evidence for case $x$ in artifact $A$,
obtained through generator $g$: either a stored record or a fresh execution
result. The instantiated property tests the obligation
\begin{equation}
  \forall x \text{ such that } P_C(x):\quad
  Q_C\!\left(x,O_C(\mathcal{R}_A^g(x))\right).
  \label{eq:claim-satisfaction}
\end{equation}
We call the outcome of checking this obligation on the supplied cases a
\emph{verdict}. We define a verdict over three case outcomes. The numeric or
uncertainty policy of $Q_C$ maps each admitted case $x$, given only its own
observations $O_C(\mathcal{R}_A^g(x))$, to
$q_C(x)\in\{\mathsf{T},\mathsf{F},\mathsf{U}\}$: the relation holds, fails,
or cannot be decided at the declared resolution. The quantifier of $Q_C$
then aggregates these outcomes.

\begin{definition}[Verdict]
\label{def:verdict}
Let $X$ be a nonempty finite set of cases that generator $g$ supplies, with
$P_C(x)$ for every $x\in X$, and let $F=\{x\in X \mid q_C(x)=\mathsf{F}\}$ and
$T=\{x\in X \mid q_C(x)=\mathsf{T}\}$. For a universal claim, which
requires every admitted case,
\begin{equation}
  V_C(X) =
  \begin{cases}
    \vviolated & \text{if } F\neq\emptyset,\\
    \vinconclusive & \text{if } F=\emptyset \text{ and } T\neq X,\\
    \vholds & \text{if } T=X.
  \end{cases}
  \label{eq:verdict}
\end{equation}
For a count claim, which requires at least $k$ of the cases that $P_C$
admits, $X$ must contain every such case, and $V_C(X)$ is \vviolated if $|X|-|F|<k$, \vholds if $|T|\geq k$, and \vinconclusive
otherwise. A \vviolated verdict carries a \emph{witness} $W\subseteq X$
with the observations $O_C(\mathcal{R}_A^g(x))$ of each $x\in W$: the
failing cases, $W=F$, for a universal claim, and all cases, $W=X$, for a
count claim.
\end{definition}

\noindent In \autoref{eq:verdict}, $V_C$ is the verdict function of claim
$C$. It is defined only on admitted and bound cases, so missing evidence or
a rejected case produces no verdict. A count claim with an unbound case
likewise produces none, because that case could still pass. For reported evidence, the rules in
\autoref{tab:scope-verdicts} refine \autoref{def:verdict}. They split
\vviolated into \vscopeflip, \vinconsistency, and \vweakened by claim form.
After every admitted case holds, they also check \emph{widened} cases that
$P_C$ excludes. A widened failure yields \vdocbound when the claim states
the exclusion and the fault verdict \voverclaim when it does not.
Changing the evidence source preserves the test only if $P_C$, $O_C$, and
$Q_C$ retain their meaning, including the metric and statistical unit.

A typed \emph{declaration} records these commitments. Its \Kind selects a
skeleton variant. Its \Domain binds the operands, scope, and policy. A
\emph{property skeleton} supplies versioned library code for that variant, and
a \emph{generator} supplies cases without rewriting that code. For C01 (\autoref{tab:worked}),
$P_C$ admits the four named datasets, $O_C$ selects the two correlations,
and $Q_C$ requires the LCA correlation to exceed the Top-1 correlation on
each dataset. The ImageNet-v2 row, which the claim explicitly excludes, is a boundary
probe,
not an admitted counterexample.

This contract has three requirements. Fixed evidence and a fixed declaration
yield the same verdict (R1). Every violation carries the observations that
decide it (R2). \aratest can reuse the declaration across cases without asking a
model to judge each case (R3). These requirements adapt the reusable
properties and counterexamples of PBT to research
records~\cite{claessen2000quickcheck,goldstein2024pbt}. They do not remove the
semantic work of translating a claim or registering a new experiment family.
Because hosted model judgments can vary even at temperature
zero~\cite{nondet2025}, \aratest lets a model propose and review
declarations but never compute a verdict from frozen evidence (R1).

\subsection{Abstracting reusable property skeletons}
\label{sec:router}

\begin{table}[t]
  \centering
  \caption{Property skeletons. Assertion forms illustrate what
    each variant preserves; they do not report reviewed whole-claim coverage
    (\autoref{sec:eval-fidelity}). We ran the invariant, threshold, and
    equivalence variants on author code (\autoref{sec:eval-execution}); two
    corpus claims have the monotonic shape, but we have not run that
    variant.}
  \label{tab:skeletons}
  \footnotesize
  \renewcommand{\arraystretch}{1.12}
  \begin{tabular}{@{}>{\raggedright\arraybackslash}p{0.21\linewidth}
    >{\raggedright\arraybackslash}p{0.29\linewidth}
    >{\raggedright\arraybackslash}p{\dimexpr0.50\linewidth-4\tabcolsep\relax}@{}}
    \toprule
    Skeleton variant & Illustrative assertion & Obligation and declaration slots \\
    \midrule
    \multicolumn{3}{@{}l}{\emph{Reported evidence: evaluated on real artifacts}} \\
    Comparative / improvement &
      A exceeds B on every named dataset. &
      Compare the subject with the reference at each in-scope coordinate.
      Bind the metric, operands, comparator, domain, and numeric policy. \\
    Consistency &
      A exceeds B on a declared number of cases. &
      Require all, or at least a declared count, of the comparisons to hold.
      Bind the reference, case set, and count. \\
    Compound consistency &
      A improves shifted-data results without degrading the original task. &
      Require each group to meet its improvement and non-degradation
      obligations. Bind the groups, operands, and requirements separately. \\
    Best-baseline consistency &
      A exceeds every eligible baseline on the required cases. &
      Check every baseline in the declared cohort, within each group.
      Bind the cohort, eligibility conditions, and required count. \\
    Numeric constraint &
      A's speedup over B is 4.7. &
      Check a value, ratio, or mean against a bound or rounded band. \\
    Reported count &
      No attack succeeds in any cell. &
      Check exact counts over a complete population. \\
    \lightmidrule
    \multicolumn{3}{@{}l}{\emph{Generative variants: fresh measurements from author code}} \\
    Metamorphic: monotonic &
      The metric follows a direction as an input increases. &
      Check an ordered perturbation sequence.
      Bind the axis, direction, metric, and tolerance. \\
    Metamorphic: invariant &
      A permitted transformation preserves the result. &
      Compare source and follow-up observations.
      Bind the transformation, metric, and tolerance. \\
    Catastrophic: threshold &
      The result stays within a declared safe range. &
      Check observations against an absolute failure threshold.
      Bind the metric, threshold, and safe side. \\
    Differential &
      A substitution moves the metric by at least a declared effect. &
      Compare paired baseline and substituted runs by their mean difference.
      Bind the factor, both values, direction, least effect, and controls. \\
    Equivalence &
      A substitution preserves a paired rate within a declared margin. &
      Check an exact interval for the paired difference.
      Bind the pairing, controls, margin, and decision policy. \\
    \lightmidrule
    \multicolumn{3}{@{}l}{\emph{Functional agreement: sampled CPU helpers}} \\
    Reference agreement &
      An implementation agrees with a reference on valid inputs. &
      Compare saved outputs under a registered arithmetic rule.
      Retain input and execution identities; sampled agreement is inconclusive. \\
    \bottomrule
\end{tabular}

\end{table}

To avoid writing a new property for each claim under test, we design a
catalog of reusable property skeletons. Each skeleton implements a recurring
relation once, such as one method outperforming another on named
datasets. A declaration then supplies the skeleton's scientific
operands. We adapt the initial
catalog from industrial PBT~\cite{goldstein2024pbt}, metamorphic
relations~\cite{chen2018metamorphic}, and property-specification
patterns~\cite{dwyer1999patterns}. A model that proposes a declaration
(\autoref{sec:schema}) selects an implemented skeleton variant rather than inventing
a new one.
The relation alone cannot express a claim's quantifier or cohort. ``Better on
every dataset'' requires each named comparison to hold, while a minimum-count
claim permits losses. ``Better than every baseline'' requires the eligible
cohort rather than one convenient reference. The declaration therefore
retains aggregation, groups, operand roles, scope, and numeric policy.
We show these obligations in \autoref{tab:skeletons}.

\begin{wrapfigure}{r}{0.45\textwidth}
  \wraptop
  \centering
  \includegraphics[width=\linewidth]{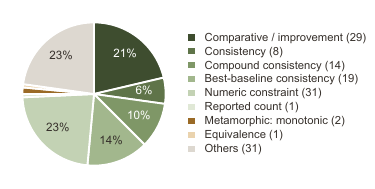}
  \caption{Classification of 136 claims from the 23 PaperBench artifacts in
    \ara-PaperBench by the skeleton variants in \autoref{tab:skeletons}. A
    claim with a declaration takes the skeleton that most of its properties use.
    This is not executable whole-claim coverage (\autoref{sec:eval-fidelity}).}
  \Description{A pie chart classifying 136 claims by skeleton variant.
    Reported-evidence variants: comparative 29, consistency 8, compound
    consistency 14, best-baseline consistency 19, numeric constraint 31, and
    reported count 1. Generative variants: monotonic 2 and equivalence 1.
    The remaining 31 claims are labeled Others. These categories do not
    establish faithful declarations, complete evidence binding, or
    whole-claim verdicts.}
  \label{fig:claim-relations}
  \wrapbottom
\end{wrapfigure}

\aratest implements six reported-evidence skeleton variants, five generative
skeleton variants, and a separate functional-agreement relation, which we list in
\autoref{tab:skeletons}. Each skeleton has a
typed parameter block, an evidence resolver, an
oracle, and an aggregation rule~\cite{qiu2026tensor,lee2026templates}. The declaration fills
the parameters; fixed library code resolves, decides, and aggregates (R1).
Adding a skeleton variant requires implementing and registering its code
under a new \Kind value. 
The generative variants compile run plans for auto-research generators.
The functional-agreement relation compares saved implementation and reference
outputs on registered inputs. The step that supplies the cases executes both
computations and retains their identities and outputs; replay does not execute
them again.
Sampled agreement remains \vinconclusive, while a mismatch retains a
counterexample (\autoref{sec:eval-fidelity}).

To guide the initial catalog design, we read the statement and
falsification criterion of 136 claims in the 23 PaperBench artifacts of
\ara-PaperBench. We then coded each claim against the current catalog
(\autoref{fig:claim-relations}). For the 41 claims that have a declaration
(\autoref{sec:eval-fidelity}), we used the skeleton that most of the
declaration's properties use. We assigned 102 claims (\qty{75}{\percent}) to
the six reported-evidence variants, including 31 numeric constraints and 29
comparative claims. Another three claims described a monotonic trend or an
equivalence within a margin, which only a generative variant expresses.
We placed the remaining 31 claims in Others: theorems, qualitative or
mechanistic analyses, correlations, growth rates, and noninferiority claims
without a stated margin.
This classification does not measure whether an implemented declaration
preserves every obligation or binds all required evidence. In
\autoref{sec:eval-fidelity}, we instead report 54 faithful, fully bound
claims out of 243 (\qty{22.2}{\percent}) after fidelity review on the corpus
that guided the catalog. That development result is not independent
scientific validation.

The catalog does not yet implement generalization, round-trip, closure,
algebraic, occurrence, or ordering properties. A \emph{router} sends each
implemented \Kind to its skeleton. A recognized kind without an
implementation receives \emph{pending-skeleton}; an unknown kind receives
\unclassified. In both cases, \aratest returns no verdict and leaves
the claim untested. Only an unknown kind counts as a
taxonomy failure.
Within one skeleton variant, different \Domain values select different obligations, such as
comparing one baseline or every eligible baseline. An experiment family
is a generator type that obtains cases (\autoref{sec:execution-backed}), not
another \Kind.

\subsection{Instantiating a skeleton from an \ara claim}
\label{sec:schema}

\begin{wrapfigure}{r}{0.48\textwidth}
  \wraptop
  \centering
  \resizebox{\linewidth}{!}{
      \includegraphics[width=\textwidth,height=0.90\textheight,keepaspectratio]{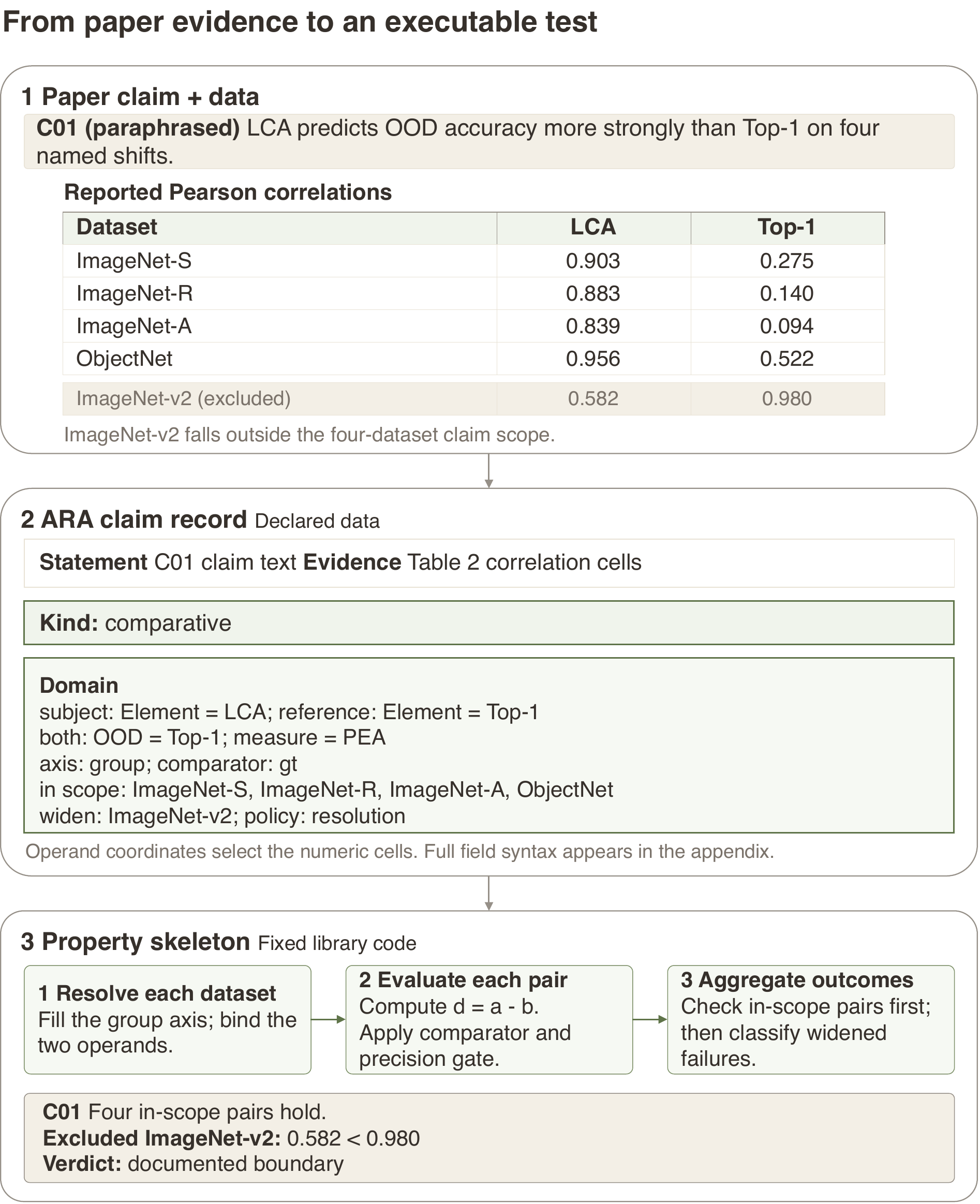}
    }
  \caption{From paper evidence to an executable test for C01. The middle panel
    highlights the added \Kind and \Domain fields, including operand coordinates
    and scope. The right panel shows how fixed library code consumes those
    fields.}
  \Description{Three panels show C01's transformation. The paper supplies a
    claim and five rows of reported Pearson correlations. An author declares
    the ARA record, or extraction proposes it. The record retains the statement
    and evidence reference and adds Kind comparative and a Domain containing
    operand coordinates, the comparison axis and comparator, four in-scope
    datasets, one widened dataset, and the resolution policy. Fixed library
    code fills the dataset axis, binds numeric pairs, applies the comparator
    and precision gate, and aggregates in-scope outcomes before classifying
    widened failures. All four in-scope pairs hold. The excluded ImageNet-v2
    pair yields a documented boundary.}
  \label{fig:ara-compilation}
  \wrapbottom
\end{wrapfigure}

A skeleton stays generic until a declaration binds it to one claim. 
We build on top of the claim record in \ara (\autoref{sec:background:ara}), 
adding two fields, \Kind and \Domain, and leaving its existing fields unchanged. 
\Kind selects a skeleton variant from the
supported vocabulary (\autoref{tab:skeletons}). The schema rejects any other
value, so \aratest leaves a claim with an invented label untested.
\Domain is a kind-specific typed declaration of what the selected skeleton variant ranges
over. It names the quantities that $O_C$ selects, the scope that $P_C$
admits, and the comparator, aggregation, and numeric or uncertainty policy
that $Q_C$ applies. Each generator reads the parts it needs
(\autoref{sec:catalog} and \autoref{sec:execution-backed}). 
We use the claim's \Statement in \ara to determine the scope in \Domain.
For C01 (\autoref{tab:worked}), \Kind is comparative, and \Domain
names the LCA and Top-1 cells, the greater-than comparator, and the four
datasets that the \Statement covers.
The transformation in \autoref{fig:ara-compilation} separates these fields from the library code that consumes them.

We intend the research agent to write \Kind and \Domain while it does the
research. \LiveResearchManager already records each claim during the
agent's research loop (\autoref{sec:background:ara}). When the agent states
a claim, it has the experiment plan, the metric, and the result tables at
hand. It can therefore declare the claim's relation, operands, and
scope in the same step, under the same provenance tags as the other \ara
fields. An author
can supply or revise these fields in the same way. The artifacts we
evaluate lack \Kind and \Domain, because the released \LiveResearchManager and
\CompilerSkill skills do not yet write declarations, and \CompilerSkill built
these artifacts from finished work. For the evaluation only, we therefore
add declarations post hoc. Given a compiled claim, its evidence, and the
catalog, an agent fills a skeleton's operands, comparator, scope, and policy
under an \aisuggested tag, or records why no skeleton expresses the claim. This step instantiates library code. It neither discovers a new
skeleton nor writes the code that computes the verdict.

Following the proposal-and-check pattern of LLM-Modulo~\cite{llmmodulo2024},
\aratest validates the schema and evidence bindings of each
proposed declaration. The agent saves its first attempt before it
sees any verdict. Binding does
not establish fidelity: a bound declaration can still select the wrong
metric or test an easier fragment of the claim. A fresh-context AI reviewer
therefore compares each declaration, or each recorded reason for skipping a
claim, with the claim and its evidence
before checking, and we report fidelity apart from verdicts
(\autoref{sec:eval-fidelity}).

\subsection{Generating cases from reported evidence}
\label{sec:catalog}

An instantiated property still needs cases to check. The reported-evidence
generator draws them from evidence that the artifact already reports: it
enumerates the table coordinates named by the declaration. For this generator, \Domain
identifies the evidence table, the two operands, the comparator, the
in-scope coordinates, the widened coordinates that the claim excludes, and
the numeric policy. \Domain also records source contracts: which part of a
cell each operand reads, which values are approximate, and which complete
population a count ranges over. The generator reads stored cells
through source-preserving readers, which keep each cell's original text,
position, printed precision, and approximation markers. It neither samples
cases nor runs new experiments. It binds no
figures, prose quantities, or cross-table relations. For C01, the dataset axis in \Domain
yields four in-scope LCA--Top-1 pairs and the excluded ImageNet-v2 pair
(\autoref{tab:worked}).

We implement this generator and its skeletons as a table auditor. The auditor
represents declaration variants as a Pydantic discriminated union, \scopespec.
The auditor's \bindfn function resolves the selected claim and table, checks the axes and
scope values, and requires each operand to identify a numeric cell uniquely.
A binding failure stops the audit instead of silently dropping a case, because
no witness can name an operand that resolves to several cells or none (R2).
The deterministic \auditscope function evaluates the bound cases and returns a
\scopefinding that records each deciding pair's coordinates, values, and
difference, so a reader can replay the verdict.

The skeleton checks the cases, using the declared comparator and displayed
precision rather than a model's judgment. For values $a$ and $b$, it computes
$d=a-b$ and a resolution gate
$\epsilon=\max(u_a,u_b)/2$, where $u_a$ and $u_b$ are the units of the
last displayed decimal places of $a$ and $b$. A pair with an approximate cell,
or with $|d|<\epsilon$, is \vinconclusive. Otherwise, the comparator decides the pair. The boundary
$|d|=\epsilon$ is decisive, while equal prints establish no order. This
gate does not test statistical significance. An effect-magnitude claim uses
its declared tolerance instead (\autoref{fig:gate-sweep}). The skeleton
then applies the declared aggregation and the rules in
\autoref{tab:scope-verdicts} from top to bottom, examining widened pairs
only after every in-scope pair passes. For widened failures, fixed
rules distinguish explicit exclusions and closed enumerations from
undeclared universal scope. Ambiguous phrasing remains inconclusive.

\begin{table}[t]
  \centering
  \caption{Reported-evidence verdict rules. The comparative relation evaluates its rows
    from top to bottom and examines widened pairs only after every in-scope
    pair passes.}
  \label{tab:scope-verdicts}
  \footnotesize
  \begin{tabular}{@{}>{\raggedright\arraybackslash}p{0.78\linewidth}l@{}}
  \toprule
  Condition & Verdict \\
  \midrule
  Comparative: any in-scope pair fails & \vscopeflip \\
  No in-scope failure, but an in-scope pair is inconclusive & \vinconclusive \\
  A widened failure has an undeclared exclusion & \voverclaim \\
  No undeclared exclusion, but a widened failure has ambiguous scope & \vinconclusive \\
  Widened failures occur only at documented exclusions & \vdocbound \\
  No widened pair fails & \vholds \\
  \midrule
  Consistency: even counting inconclusive pairs as passes cannot meet the required count & \vinconsistency \\
  Compound: all comparison requirements hold, but a declared non-degradation requirement fails & \vweakened \\
  \bottomrule
\end{tabular}

\end{table}

\subsection{Generating cases with auto-research}
\label{sec:execution-backed}

\begin{figure*}[t]
  \centering
  \resizebox{0.88\textwidth}{!}{\begin{tikzpicture}[
  font=\sffamily\scriptsize,
  box/.style={draw=oasiswarmgray, rounded corners=2pt, align=center,
    text width=6.2cm, inner sep=4pt, minimum height=0.92cm, fill=white},
  fixed/.style={box, draw=oasisolive, fill=oasisolivebg},
  flow/.style={-{Stealth[length=1.8mm]}, thick, draw=oasischarcoal}
]
  \def\arrowlen{0.35cm}
  \node[fixed, text width=13.6cm] (contract) at (7,0)
    {\textbf{\footnotesize Fixed claim-testing contract}\\[2pt]
    $P_C$: admitted scope \qquad $O_C$: required observations
    \qquad $Q_C$: relation, aggregation, and decision policy};
  \node[box, anchor=north] (reported) at ($(contract.south -| 3.3,0) + (0,-\arrowlen)$)
    {\textbf{\footnotesize Reported-evidence generator}\\[2pt]
    Read stored results for the declared comparisons.};
  \node[box, anchor=north] (proposal) at ($(contract.south -| 10.7,0) + (0,-\arrowlen)$)
    {\textbf{\footnotesize Auto-research generators}\\[2pt]
    Take the experiment families bound to the claim in the \ara.};
  \node[box, anchor=north] (enumerate) at ($(reported.south) + (0,-\arrowlen)$)
    {\textbf{\footnotesize Enumerate cases}\\[2pt]
    Use the cases already reported.};
  \node[box, anchor=north] (admission) at ($(proposal.south) + (0,-\arrowlen)$)
    {\textbf{\footnotesize Admission check}\\[2pt]
    The experiment-generator agent declines substitutions that the claim fixes.};
  \node[box, anchor=north] (stored) at ($(enumerate.south) + (0,-\arrowlen)$)
    {\textbf{\footnotesize Stored observations}\\[2pt]
    Retain the evidence for each case.};
  \node[box, anchor=north] (execute) at ($(admission.south) + (0,-\arrowlen)$)
    {\textbf{\footnotesize Experiment-generator agent}\\[2pt]
    Run additional experiments on the artifacts and report.};
  \node[fixed, text width=13.6cm, anchor=north] (check) at ($(stored.south -| 7,0) + (0,-\arrowlen)$)
    {\textbf{\footnotesize Property check outside every generator}\\[2pt]
    Select observations with $O_C$, then evaluate $Q_C$ and aggregate.\\
    Retain deciding evidence for replay and source adjudication.};
  \draw[flow] (contract.south -| reported.north) -- (reported.north);
  \draw[flow] (contract.south -| proposal.north) -- (proposal.north);
  \draw[flow] (reported.south) -- (enumerate.north);
  \draw[flow] (enumerate.south) -- (stored.north);
  \draw[flow] (proposal.south) -- (admission.north);
  \draw[flow] (admission.south) -- (execute.north);
  \draw[flow] (stored.south) -- (check.north -| stored.south);
  \draw[flow] (execute.south) -- (check.north -| execute.south);
\end{tikzpicture}}
  \caption{Two generators supply cases to one fixed claim property: one reads
    stored observations, and the other runs registered experiment families.}
  \Description{A fixed claim contract feeds two paths. One reads reported
    evidence. The other takes the experiment families bound to the claim
    and dispatches the experiment-generator agent, which declines
    configuration substitutions that the claim fixes and runs the remaining
    instances. Both paths supply observations to the same property check.}
  \label{fig:two-generators}
\end{figure*}
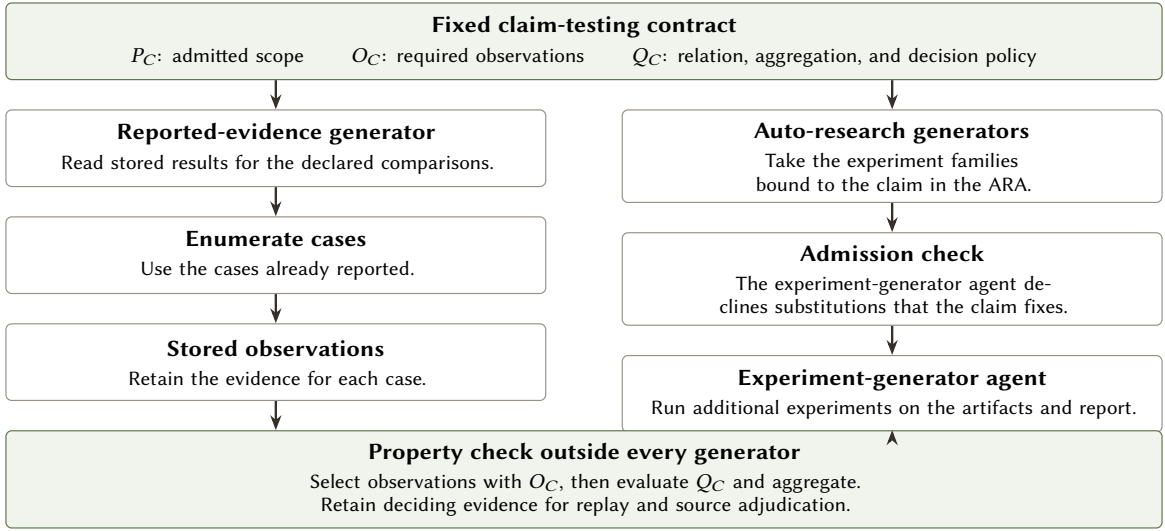

Reported evidence can test a claim only where the authors measured it.
To obtain fresh measurements instead of reading stored ones, we delegate
experiments to an AI agent that plans and runs
them~\cite{kon2025curie}. We call this agent the
\emph{experiment-generator agent}. We neither hard-code the
experiments nor ask the agent to invent them. Instead, each generator
describes one \emph{experiment family}, a way to obtain related cases such as
resampling random seeds or permuting row order while the claim's task stays
fixed. When such a generator draws cases, it gives the agent the claim's \ara
or the paper's code repository together with the family description. The
agent runs the experiments and returns observations of the case type that the
property expects. We call these generators \emph{auto-research generators}.
The property still decides each case, but a failing case on this path
serves as feedback rather than a refutation (\autoref{fig:two-generators}).

For an auto-research generator, \Domain supplies the controls and uncertainty
policy that each family needs. The family description instructs the agent not to
substitute a metric or method variant, drop a baseline, or loosen the
decision rule after observing a failure. Each case must also keep the unit
over which $Q_C$ quantifies. Matched runs retain the declared controls, and a
seed batch retains its statistical unit and aggregation, so one adverse seed
does not refute a claim about a mean. Comparing a run with a published value
differs from comparing it with a baseline and requires a separate
declaration. For C01 (\autoref{tab:worked}), fresh evidence would still compare LCA and Top-1
predictive correlations on the four named datasets.

\subsubsection{Generators and experiment families}
\label{sec:claim-scoped-generation}

Our Python library enforces the case type through one protocol
(\autoref{sec:generator-interface}). A
\texttt{Generator} names its case type \texttt{C}, such as a batch of seeded
runs or a matched pair of runs. Each call to its \texttt{generate} method
returns a batch of cases of that type with its coverage: how many cases the
generator meant to supply and why any are missing. The property must take
the same type, and \aratest withholds the verdict on a batch whose
coverage is incomplete. The reported-evidence generator
(\autoref{sec:catalog}) is a \texttt{Generator} whose cases are pairs of
stored table cells. An auto-research generator runs one experiment family
for one claim. Each call dispatches the experiment-generator agent with the
artifact, the declaration, and the family's description, and the generator
accepts only returned cases of type \texttt{C}. \aratest fixes this request
and the report format, but not the agent. An agent registers under the
\texttt{aratest.agents} entry-point group, so any agent that accepts the
request and reports cases can serve. Our \dissclaimer plugin registers one
such agent, which runs each experiment in an OpenHands
workspace~\cite{wang2025openhands}.

A valid declaration does not guarantee a valid run. The agent implements
each experiment itself and may change more than the family allows, for
example by substituting a metric, dropping a baseline, or skipping a
control. Agent-designed probes can also include cases that human experts
reject~\cite{bai2026story}. We therefore do not treat a failing case from an
auto-research generator as a refutation of the claim. \aratest reports
it as \emph{feedback}: a candidate violation together with the agent's code,
logs, and observations. The feedback suggests the next experiment to run or
a part of the artifact to revise, and the user decides whether the
failing case reveals a claim defect or a flawed run (\autoref{sec:soundness}).

By default, the library provides four family types:
\begin{itemize*}
\item \texttt{SeedResampling}: reruns the experiment under fresh random
  seeds.
\item \texttt{OrderPermutation}: reorders data rows while keeping groups,
  pairs, and partitions intact.
\item \texttt{NullCalibration}: runs the method on data drawn under a null
  hypothesis to measure its error rate.
\item \texttt{ConfigSubstitution}: replaces one released configuration value
  with a declared alternative, such as a value stated in the paper, once the
  experiment-generator agent admits the substitution.
\end{itemize*}
The first three types vary randomness that claims rarely commit to, so they
need no admission check. A claim may instead fix a configuration value, as a
claim stated at a learning rate of 0.1 fixes the learning rate. Before it
runs a substitution, the experiment-generator agent therefore reads the \ara
and decides whether the claim leaves the substituted value open. Users can
extend the library with their own family types. Each family type is data:
an instruction, a parameters model, and a case type. Users register a new type under the \texttt{aratest.families}
entry-point group. A family type is generic, and an instance
binds it to one artifact and claim. The six property-based tests in our
evaluation draw cases from six instances of these four types
(\autoref{sec:eval-execution}).

\subsubsection{Declaring and running generators}
\label{sec:family-composition}

In our design, the research agent chooses a claim's auto-research generators
when it states the claim, as it does for the claim's declaration
(\autoref{sec:schema}). It
names the family types that apply and fills their parameters, such as the
seed count, the permuted partition, or the substituted value. It records
this binding with the claim in the \ara under an \aisuggested tag, much as a
PBT user annotates a test with its generators. An author can supply or
revise the binding in the same way. \aratest does not choose generators.
It reads the recorded binding. For a \texttt{ConfigSubstitution} instance,
the experiment-generator agent declines the substitution and returns no
cases if the claim fixes the substituted value.

\begin{figure*}[t]
  \centering
  \includegraphics[width=\textwidth]{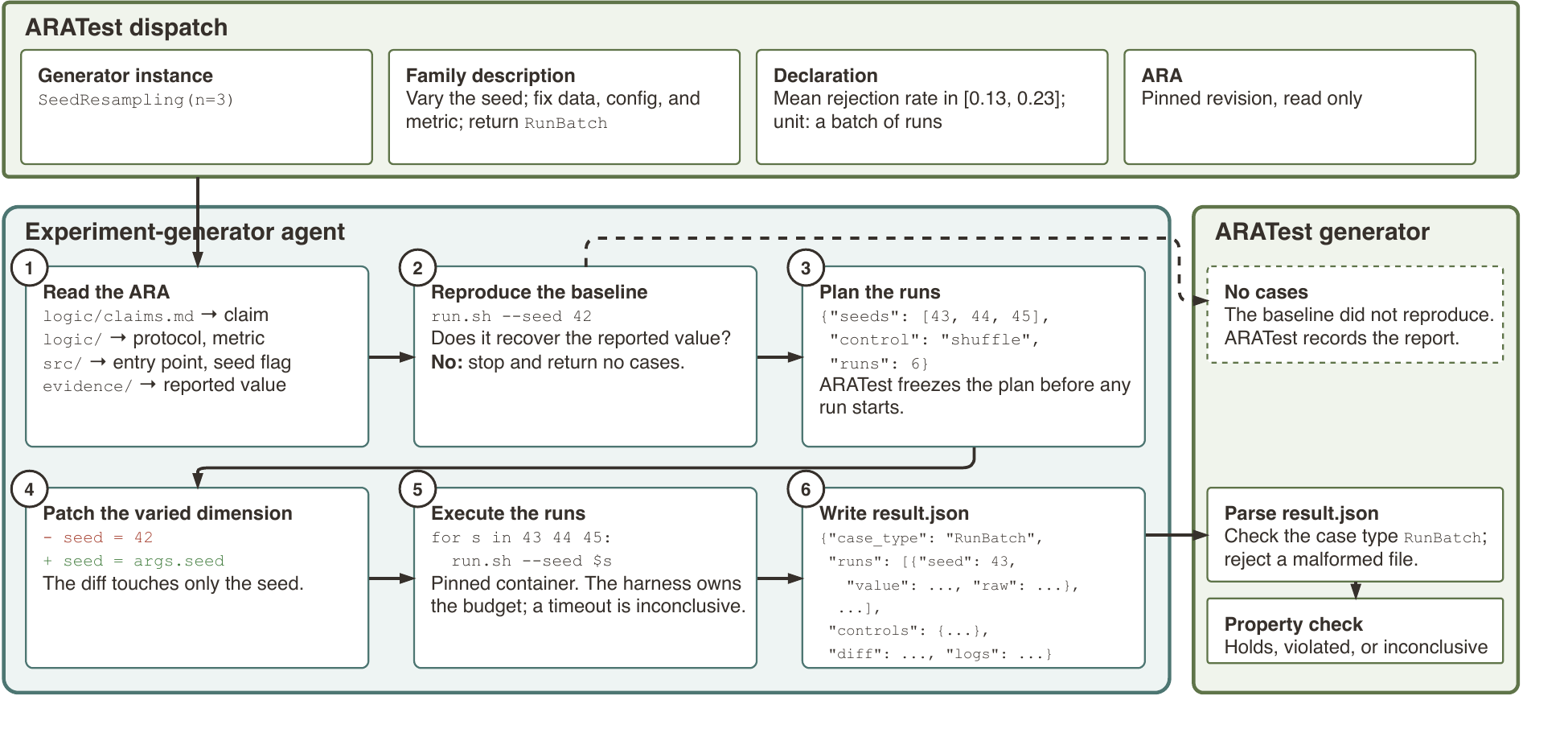}
  \vspace{-3em}
  \caption{Workflow of the experiment-generator agent, with a
    \texttt{SeedResampling} instance for an \mmdfuse claim as an example.
    The numbered steps trace one agent run.}
  \Description{Three lanes. A top lane, ARATest dispatch, holds four
    inputs: the generator instance SeedResampling with three seeds, the
    family description (vary the seed; fix data, config, and metric; return
    a RunBatch), the MMD-FUSE declaration (mean rejection rate in 0.13 to
    0.23, unit a batch of runs), and the ARA at a pinned, read-only revision.
    The middle lane shows six numbered agent steps: read claims, plans,
    source, and evidence from the ARA; rerun the baseline and stop with no
    cases if it does not reproduce; plan the seeds and a shuffle control,
    which ARATest freezes; patch only the seed; execute the runs in a pinned
    container; and write result.json with the runs, controls, diff, and
    logs. A dashed arrow from step two leads to a No cases box in the right
    lane. In the right lane, ARATest's generator parses result.json
    against the RunBatch case type and passes it to the property check,
    which returns holds, violated, or inconclusive.}
  \label{fig:experiment-generator-agent}
\end{figure*}

For each remaining instance, \aratest dispatches the
experiment-generator agent with the family description, the declaration, and
the \ara at a pinned revision (\autoref{fig:experiment-generator-agent}). The
agent first reads the \ara \stepmark{1}: the claim in \texttt{logic/claims.md}, the
experiment plan behind its \Proof link, the entry point and configuration in
\texttt{src/}, and the reported value in \texttt{evidence/}. It then reruns
the released code at the original setting \stepmark{2}. If that run does not recover the
reported value, the agent stops and returns no cases. Otherwise, it plans the
runs that the family calls for, where a run executes the released code once at
one setting. \aratest freezes the plan before any run starts
\stepmark{3}. The agent patches only the dimension that the family varies
\stepmark{4}, executes the runs in a pinned container under a fixed time and
compute budget \stepmark{5}, and writes a JSON file with the measurements, controls, code diff, and logs \stepmark{6}. 
\aratest's generator parses the file into
cases of the family's case type, rejects a malformed file, and passes the
cases to the property. The agent chooses how to realize the experiment and
decides whether to admit a substitution, but it does not write the property.

The library combines generators with two combinators. Each call to
\texttt{one\_of} draws a batch from the next of several generators in turn,
and \texttt{frequency} weights that choice. Both require their generators to
share a case type, so a combination draws every case in one
statistical unit. For \mmdfuse~\cite{biggs2023mmdfuse}, the seed and
row-order instances both yield batches of repeated two-sample tests, so one
generator can draw from either. The null-calibration and
configuration-substitution instances test other claims with their own declarations, whose
properties take other case types. \dissclaimer's continuous testing service
schedules each family binding separately and does not combine generators.

\aratest retains each instance's frozen plan and returned JSON file, so a
reader can recompute the verdict without rerunning the research code.
For that replay to test the same claim, retained cases keep the
claim's unit: a seed family preserves the randomness over which its claim
quantifies, and a reduced counterexample must retain the original scope,
pairing, and controls. Only completed runs reach
the property. \aratest records every incomplete run with its reason, so
a reader can see which runs produced no result. 

\subsection{What a verdict does and does not establish}
\label{sec:soundness}

A verdict decides the instantiated property on the recorded evidence, not
the source claim. The record keeps the witness: the deciding cells, or the
agent's frozen design, run record, and observations. Because each case
outcome depends on that case alone (\autoref{def:verdict}), the witness
suffices to replay a violation.

\begin{claimproposition}[Replayable violations]
\label{thm:witness}
Let $V_C(X)=\vviolated$ with witness $W$. Then
\begin{enumerate*}[label=(\roman*)]
\item $V_C(W)=\vviolated$, computed from the observations recorded in $W$
  alone; and
\item for a universal claim, $V_C(X')=\vviolated$ for every finite set
  $X'\supseteq W$ of admitted cases.
\end{enumerate*}
\end{claimproposition}

\begin{proof}
By \autoref{def:verdict}, $q_C(x)$ depends only on $x$ and its recorded
observations, and fixed library code computes it identically on every run
(R1). For a universal claim, $W=F$ is nonempty and $q_C(w)=\mathsf{F}$ for
every $w\in W$, so every $w\in W$ belongs to the failing set of every
$X'\supseteq W$, including $W$. For a count claim, $W=X$, so
$V_C(W)=V_C(X)$.
\end{proof}

\noindent For a universal claim, a \vholds verdict has neither guarantee,
because one added failing case turns it into \vviolated.
The replay guarantee (\autoref{thm:witness}) does not cover \voverclaim, which
rests on widened cases outside $X$. Replaying it also requires every admitted
case, because the rules examine widened pairs only after all admitted pairs
pass. Replay checks only the step from evidence to verdict. Attributing a
fault to the source work requires checking that the \CompilerSkill skill did
not paraphrase the claim or mislabel a table, and that the declaration bound
the right metric. The record keeps the \EvidenceBasis and the declaration so
that a reader can make this check.

\aratest reports violations as warnings and leaves the decision about a
defect to its user, much as GCC lets its user promote warnings to errors with
\texttt{-Werror}. By default, only the four fault verdicts in
\autoref{tab:scope-verdicts} (\voverclaim, \vscopeflip, \vinconsistency, and
\vweakened) recommend changing \Status to \strefuted. A stricter user can
treat every violation as a defect, even on the auto-research path. A
statistical rejection holds only under the declared uncertainty policy. A pass
corroborates only the obligations and cases checked, so
\aratest never promotes the claim.

On the auto-research path, a reader must also use the code diff and controls
to check that the agent's run realized the family. This check matters for a
pass too, because a run that left the seed unchanged would return \vholds
without testing the claim. By default, a violation on this path is
feedback, not a \Status recommendation
(\autoref{sec:claim-scoped-generation}). A baseline
that does not reproduce yields no cases, a finding about the released code
rather than the claim. An \vinconclusive result neither supports nor weakens
the claim.

\section{Using \aratest}
\label{sec:usage}

We now show what a user writes and what \aratest returns. On two claims
from \ara-PaperBench, we walk through the three steps of integration
testing: declaring a claim, testing it, and replaying a violation. We then
bind an experiment family to a claim for continuous testing and add a new
family to the library. We ran every command whose output we show, and
\texttt{aratest validate} accepts every spec we show. A listing marks each
place where we shortened the text.

\subsection{Declaring a claim}
\label{sec:usage-declare}

A user declares the tests of an \ara in one file, \texttt{aratest/spec.yaml},
inside the artifact. The spec holds no code and repeats no reported number.
It names table coordinates, and \aratest reads the numbers from the
evidence files. In \autoref{fig:c01-spec}, we declare the comparison of
claim C01 in \emph{LCA-on-the-Line} (\autoref{tab:worked}).

\begin{figure}[t]
  \centering
  \input{listings/c01-spec}
  \caption{The spec that declares the comparison of C01 in
    \emph{LCA-on-the-Line}. It names the evidence table and the cells to
    compare, but no reported value.}
  \label{fig:c01-spec}
\end{figure}

The spec has two parts. The \texttt{evidence} block tells \aratest how to
read each evidence table into measurements. The correlation table lists
one predictor per row and labels each column with a dataset and a
statistic, such as \texttt{ImgN-S PEA}. The block therefore takes each
cell's metric from the statistic, renames the row axis to
\texttt{predictor} and the column group to \texttt{dataset}, and maps the
row labels to \texttt{LCA} and \texttt{Top1}. The \texttt{claims} block
binds C01 to the comparative skeleton, whose fields fill the contract of
\autoref{sec:contract}. The fields \texttt{metric}, \texttt{subject}, and
\texttt{reference} define the observation map $O_C$, and
\texttt{in\_scope} defines the domain predicate $P_C$. The relation $Q_C$
applies \texttt{comparator} on every in-scope dataset. The \texttt{widen}
list names ImageNet-v2, which the claim excludes, as a widened case.
Before testing, \texttt{aratest validate} lists every property that would
not bind and why, such as an operand that resolves to no cell.

We wrote this declaration for the walkthrough, and it is not one of the
RQ1 declarations. It covers only the comparison in \autoref{tab:worked}.
The compiled C01 also requires LCA's $R^2$ and Pearson correlation to
exceed 0.7 on each dataset, which a complete declaration would check with
numeric constraints.

\subsection{Testing the claim}
\label{sec:usage-test}

\texttt{aratest test} reads the spec, binds each property, builds its cases
from the evidence, and prints a verdict for each claim
(\autoref{fig:c01-test}). C01 holds because LCA's correlation exceeds
Top-1's on all four datasets. The refinement \vdocbound reports the
widened case. ImageNet-v2 reverses the comparison, but the claim's
statement enumerates the four datasets, so \aratest classifies the
reversal as a boundary of the claim rather than a fault
(\autoref{tab:scope-verdicts}). The exit code summarizes the run for
scripts and agents: 0 when every declared claim holds, 1 when a claim is
violated, and 4 when a claim stays undecided. With \texttt{-{}-format json},
the command prints the same result as one JSON document.

\begin{figure}[t]
  \centering
  \input{listings/c01-test}
  \caption{Testing C01 against the reported correlations. The claim holds
    on its four datasets, and the ImageNet-v2 reversal marks a documented
    boundary.}
  \label{fig:c01-test}
\end{figure}

\subsection{Replaying a violation}
\label{sec:usage-replay}

A violated claim leaves a record that replays without the artifact. In
\autoref{fig:apt-replay}, we test claim C02 of
\apt~\cite{zhao2024apt}, one of the two paper errors in
\autoref{sec:eval-fidelity}, under its RQ1 declaration. The claim states
that \apt trains 8.4$\times$ faster than LoRA+Prune on RoBERTa and
8.2$\times$ faster on T5. The declaration checks each speedup with a
\texttt{numeric\_constraint} property: the ratio of two training-time cells
must round to the stated value at one decimal place. For RoBERTa, the
cells \num{5128.3} and \num{592.1} bound the ratio to $[8.660, 8.662]$
within their printed precision, outside the band $[8.35, 8.45]$ that a
stated 8.4 permits. Both properties are violated, so C02 is violated, and
\aratest saves a check record under \texttt{.aratest/records/}, named by
the digest of its content.

\begin{figure}[t]
  \centering
  \input{listings/apt-replay}
  \caption{Testing C02 of \apt and replaying its record after moving the
    \ara away. We shortened the record digest.}
  \label{fig:apt-replay}
\end{figure}

The record holds every input of the check: the declaration and its
provenance, the relation's kind, version, and source digest, the claim's
statement, and each bound measurement with its value, printed precision,
and source line. \texttt{aratest replay} recomputes each verdict from the
record alone and prints the recorded and recomputed outcomes together.
We moved the \ara out of the working directory before replaying, so the
replay read no evidence file. The witness in the record thus reproduces
the violation, as \autoref{thm:witness} states. Replay warns when the
relation's source has changed since the record was saved, and it stops
when the relation's version has changed.

\subsection{Binding experiment families}
\label{sec:usage-families}

Reported evidence cannot test behavior that the authors did not measure.
For such a claim, a user binds experiment families in the same spec
(\autoref{fig:family-binding}). We show the declaration of \mmdfuse claim
C12 from our campaign (\autoref{sec:eval-setup}). Its \texttt{invariant}
property requires the mean rejection rate over seeds to stay in
$[0.13, 0.23]$, a band that we chose for the study, and requires a
label-permutation control to lower the rate by at least 0.05. The
\texttt{families} block binds \texttt{seed\_resampling}, which varies only
the \texttt{jax\_seed} stream, and asks the agent to run the control.

\begin{figure}[t]
  \centering
  \input{listings/family-binding}
  \caption{The declaration of \mmdfuse claim C12 with a
    \texttt{seed\_resampling} binding. We shortened the control's
    instruction to the agent.}
  \label{fig:family-binding}
\end{figure}

\texttt{aratest test} runs no experiment. It reports the
\texttt{invariant} property as \texttt{no verdict (missing\_evidence)},
because the property needs cases of type \texttt{aratest.run-batch}, and
exits with code 4. With \dissclaimer installed,
\texttt{aratest continuous start} runs the bound families as a campaign. Its
options name the agent, here \dissclaimer's \texttt{openhands}, a budget
for each run, and a budget for the campaign. For each binding, the service
dispatches the experiment-generator agent
(\autoref{fig:experiment-generator-agent}) and checks the property on the
batch that the agent returns. \texttt{aratest continuous status} reports
the budget spent and the runs by status, and
\texttt{aratest continuous findings} lists each violation with its record
and the \texttt{aratest replay} command that rechecks it without
\dissclaimer. We do not reproduce campaign output here; we report our
campaign's batches in \autoref{tab:continuous-campaign}.

\subsection{Extending the library}
\label{sec:usage-extend}

\aratest loads its relations, experiment families, agents, and commands
from Python entry points, so a user extends it by installing a package
rather than editing \aratest. In \autoref{fig:family-define}, we define a
family that reruns training under several seeds. A family is data: an
instruction for the agent, a parameters model that a binding fills, and
the case model that the agent's report must match.
\texttt{ExperimentFamily.define} derives JSON schemas for the parameters
and the cases from the two Pydantic models. The package registers
\texttt{SCORE\_SEEDS} under the \texttt{aratest.families} entry-point group
in its \texttt{pyproject.toml}, and a spec can then bind
\texttt{score\_seeds} as in \autoref{fig:family-binding}.

\begin{figure}[t]
  \centering
  \input{listings/family-define}
  \caption{A user-defined experiment family. The case model fixes the type
    of every case that the family supplies.}
  \label{fig:family-define}
\end{figure}

A new relation registers under \texttt{aratest.relations}. It declares a
domain model, which validates a declaration's fields, and a case model,
and it implements four methods. \texttt{bind} resolves what the relation
needs from the artifact, \texttt{observe} reads one case, \texttt{fold}
decides from all observations, and \texttt{refine} labels the decision.
Because \texttt{bind} records a digest of the relation's source, replay
detects a relation that changed after a record was saved. Until a package
registers a relation for a \Kind, \aratest leaves claims of that kind
untested (\autoref{sec:router}). An agent registers under
\texttt{aratest.agents}, as \dissclaimer registers its OpenHands agent
under the name \texttt{openhands}. A package can also add a top-level
command under \texttt{aratest.commands}, and \dissclaimer adds
\texttt{aratest continuous} this way.

\section{Evaluation}
\label{sec:evaluation}

We evaluate \aratest on three questions: declaration fidelity on
artifacts (RQ1), seeded-fault detection on common evidence (RQ2),
and scope-preserving experiments on author code (RQ3).

\subsection{Experimental setup}
\label{sec:eval-setup}

Every subject enters \aratest as an \ara artifact. We drew the subjects
from the \ara-PaperBench corpus~\cite{liu2026ara,paperbench2025}, which holds
33 \ara artifacts: 23 from PaperBench papers, five from RE-Bench
tasks~\cite{wijk2025rebench}, one from the NanoGPT speedrun, three from other
papers, and our \mmdfuse artifact. We compiled
the two subjects of the reviewed RQ3 tests, \mmdfuse and \simformer, from
their papers and released code with the \CompilerSkill skill
(\autoref{tab:subjects}). We run two groups of
experiments, one for each testing mode (\autoref{sec:approach}). The main
coverage and detection experiments (RQ1 and RQ2) evaluate integration
testing on reported evidence; apart from the author CPU helpers that RQ1 and
RQ2 compare against references, they execute no author code.
The extended experiments (RQ3) evaluate continuous testing on author code,
which the \dissclaimer plugin runs.

\needspace{6\baselineskip}
\subsubsection{Integration testing on reported evidence}
\begin{wraptable}{r}{0.40\columnwidth}
  \wraptop
  \caption{Evaluation subjects. All are \ara artifacts compiled from published papers.}
  \wrapcapskip
  \label{tab:subjects}
  \footnotesize
  \begin{tabular}{@{}>{\raggedright\arraybackslash}p{\dimexpr0.66\linewidth-\tabcolsep\relax}
    >{\raggedright\arraybackslash}p{\dimexpr0.34\linewidth-\tabcolsep\relax}@{}}
    \toprule
    Subject and topic & Test units \\
    \midrule
    \multicolumn{2}{@{}p{\linewidth}@{}}{\emph{Existing artifacts, reported evidence: RQ1}} \\
    33 \ara-PaperBench artifacts\newline ML, R\&D, and systems papers & 243 claims \\
    \midrule
    \multicolumn{2}{@{}p{\linewidth}@{}}{\emph{Seeded faults, common evidence: RQ2}} \\
    \sema~\cite{wang2025sema}\newline class-incremental learning & 288 faults \\
    \cfgartifact~\cite{sanchez2023stay}\newline guidance decoding & 265 faults \\
    RE-Bench Triton~\cite{wijk2025rebench}\newline cumulative-scan helper & 3 faults \\
    \midrule
    \multicolumn{2}{@{}p{\linewidth}@{}}{\emph{Unseeded, executed author code: RQ3}} \\
    \mmdfuse~\cite{biggs2023mmdfuse}\newline kernel two-sample testing & 4 tests, 55 runs \\
    \simformer~\cite{gloeckler2024allinone}\newline simulation-based inference & 2 tests, 9 runs \\
    \midrule
    \multicolumn{2}{@{}p{\linewidth}@{}}{\emph{Continuous testing campaign, author code: RQ3}} \\
    \mmdfuse C12 & 5 batches \\
    \pinn~\cite{rathore2024pinn}\newline physics-informed networks & 1 batch \\
    \smm~\cite{cai2024smm}\newline visual reprogramming & 1 batch \\
    \bottomrule
\end{tabular}

  \wrapbottom
\end{wraptable}

For RQ1, we studied all 243 claims in the 33 artifacts after withholding
existing declarations. An agent, Codex with \texttt{gpt-6-astra}, declared
each claim or recorded why it skipped the claim, under the protocol of
\autoref{sec:schema}. We kept source bytes fixed within each
comparison.

For RQ2, we seeded faults into \sema and \cfgartifact.
Four mutation operators reverse or degrade the claimed comparison. A value
swap exchanges the treatment and baseline cells, a single-coordinate
counterexample moves one cell by one displayed unit, and two shift operators
move the baseline or the treatment past the flip point by
$k \in \{1, 2, 5, 10\}$ displayed units or by the whole pristine margin. We
kept a mutant only if mechanical checks confirmed that exactly the intended
cells changed and the claim's truth value flipped. This yielded 553 mutants.
Automated review judged all 553 semantically valid and labeled 491 of
them \lplausible (\autoref{sec:eval-threats}). We exposed 193 mutants during
development and held out the 360 shift mutants at $k \in \{2, 5, 10\}$ and
the whole-margin setting, so the holdout is over mutations, not papers. We
paired each mutant with its unedited original and added 18 narrower census
controls that no operator edited. Before any method predicted, a separate
validator labeled every comparison with exact rational arithmetic against
the full declared property: six clauses of six comparisons for \sema and
eight clauses of three for \cfgartifact. The validator counts a fault as exposed only
when a comparison that the mutation affects passes in the original and
fails after the mutation; all 553 faults are exposed. The 553
mutants, their 553 originals, and the 18 controls form \num{1124} records.
We gave each checking method one \emph{packet} per record; a packet holds the
property and the observations to check. Because records with identical
inputs share a packet, the \num{1124} records collapse to 504 packets:
485 violate the property and 19 hold. The 19 holding packets are the
valid controls. The 265 originals that already violate another required
comparison share one violated packet and so are not controls.

We also seeded faults into author code. We altered the cumulative-scan helper of
the RE-Bench Triton artifact in three ways: the positive test \texttt{x > 0}
became \texttt{x >= 0} or \texttt{x > 1}, or the parity test was inverted.
An equivalent predicate, \texttt{0 < x}, served as a control. A
network-isolated sandbox ran the original, the three variants, and the
equivalent twice each on the 12 registered inputs that RQ1 uses for this
helper (\autoref{sec:eval-fidelity}). Its 120 evaluations form four packets, each
checked by a sampled agreement property over 61 comparisons with a separate
sequential reference. The three variants violate it, and the shared
original-and-equivalent packet is \vinconclusive, because agreement on
sampled inputs cannot establish correctness.

We compare the deterministic checker with a direct LLM judge and a
calculator-assisted LLM judge. All three receive the same packet (\autoref{tab:methods}): the
plain-language property, its structured clauses, and the observations with
comparison identities, units, uncertainty, and completeness limits.
Recorded packets also carry the raw table and claim text. Fault labels,
mutation locations, and reference verdicts stay hidden. The deterministic
checker applies \aratest's relations once. Both LLM judges use
\texttt{gpt-6-luna} with medium reasoning effort and provider sampling
defaults, under a \num{20480}-byte input limit for recorded packets and
\num{1536} output tokens per request. The calculator-assisted judge may also
call an exact rational-arithmetic tool, but it cannot run code, read files,
or invoke the checker. We score only the first of each judge's five
fresh-context trajectories and keep a failed first trajectory in the
denominator. A method detects a fault when it declares a violation and
cites a valid witness: a comparison that the fault affects, with correct
operands, arithmetic, and clause. Citing a
different defect does not detect the injected fault.

\begin{table}[t]
  \caption{Methods compared in RQ2. All three receive the same
    packet; the judges share one prompt and response schema
    (\autoref{sec:judge-prompt}).}
  \label{tab:methods}
  \footnotesize
  \begin{tabular}{@{}p{0.25\linewidth}p{0.22\linewidth}p{0.29\linewidth}p{0.14\linewidth}@{}}
    \toprule
    Method & Decision procedure & Tools & Runs per packet \\
    \midrule
    Deterministic checker & \aratest relations & None & 1 \\
    Direct judge & \texttt{gpt-6-luna} & None & 5, first scored \\
    Calculator-assisted judge & \texttt{gpt-6-luna} & Exact rational arithmetic, up to 4 rounds & 5, first scored \\
    \bottomrule
\end{tabular}

\end{table}

\subsubsection{Continuous testing on artifact code}

For
each candidate subject, we used the \CompilerSkill skill to compile an \ara
artifact from the paper and its released author code. We kept an artifact
only when its author code ran in a pinned container and reproduced a reported
value without reading that value. Two of six candidate artifacts qualified: \mmdfuse~\cite{biggs2023mmdfuse}, which measures rejection
frequency over repeated two-sample tests, and
\simformer~\cite{gloeckler2024allinone}, which measures classifier two-sample
accuracy (C2ST, chance level \num{0.5}). The others missed a frozen interval,
required unavailable hardware, exceeded the time limit, or lacked a license
grant.
Neither subject carries an injected fault, but we saw both during
development.

We ran six property-based tests on the unmodified author code.
Each test pairs one claim's property with one auto-research generator, and
\dissclaimer's experiment-generator agent runs the generator's experiments. The
generators instantiate all four family types on \mmdfuse, and only
\texttt{SeedResampling} and \texttt{OrderPermutation} on \simformer
(\autoref{tab:execution-families}). The \mmdfuse seed and row-order tests
share one property and differ only in their generator.
Neither artifact records a generator binding, so we supplied the binding in
place of the research agent. A fresh-context AI reviewer checked each design
after the experiment-generator agent produced it and before we froze its
target quantity, interventions, controls, and decision rule. We also
evaluated one paired reported/fresh
property on \mmdfuse's released implementations, separately from the six
tests.

Each run executed in a Docker image pinned by digest under a one-hour limit on
one 80-core Linux server. \mmdfuse tests received 8 CPU cores and
\qty{16}{\gibi\byte} of memory; \simformer tests received 4 cores,
\qty{32}{\gibi\byte}, and one NVIDIA Quadro RTX 8000. We tested the generative
catalog only on synthetic data with known noise scale $\sigma$ and decision
band $h=\max(0.05,3\sigma)$, so the real-code tests used their own
registered uncertainty rules.

We then ran \dissclaimer's continuous testing service without human
review. We screened six candidate targets from the RQ1 corpus and admitted
three targets whose unchanged author code and data we could obtain and whose
runs fit on one server: \mmdfuse claim C12, \pinn~\cite{rathore2024pinn} claim
C06, and \smm~\cite{cai2024smm} (three subclaims of claims C03 and C05). The other three
lack public author code or need days of ImageNet training or multi-GPU
LLaMA-scale pruning per run. An authoring agent on \texttt{claude-opus-5-5} wrote each
declaration under the protocol of \autoref{sec:schema}. Before any run, a
fresh-context AI reviewer checked each declaration against its source, and
the authoring agent revised any declaration that the reviewer rejected. The \mmdfuse C12 band is our study assertion, not the paper's. Each
target and model pair ran as its own campaign, in which the service
dispatched the experiment-generator agent on \texttt{gpt-6-luna} or
\texttt{glm-5.3-flash}. Each run received offline package and data caches, plus two Quadro RTX
8000 GPUs and about \qty{20}{\hour} for \pinn and \smm, or no GPU and
about \qty{2.7}{\hour} for \mmdfuse. A violation would count as a finding only after a
fresh-context AI review against the source, a replay with the
deterministic checker alone, and a confirmation run with new random draws. We report every batch
that the service adopted within the campaign window.

\subsection{RQ1: Declaration fidelity on existing artifacts}
\label{sec:eval-fidelity}

In RQ1, we ask which complete claims an agent can faithfully declare and
\aratest can check. We designed the catalog and readers on these 243
claims (\autoref{sec:catalog}), so the RQ1 results bound what the current
catalog reaches and do not estimate coverage on new papers. A claim is \emph{faithful and
bound} when the reviewer accepts its whole declaration and every required
property binds to evidence. It is \emph{decided} when that bound declaration
holds or is violated. Every rate divides by all 243 claims, including skips.

\begin{table}[h]
  \centering
  \caption{RQ1 ablation on the 243 claims in 33 artifacts. Each row
    removes one more component from the row above, so the last row runs
    the base relation catalog alone. Restatements are decisions that rest
    on reported summaries alone. $^{\dagger}$Without the RE-Bench
    source-format loader, these rows exclude two RE-Bench claims that the
    rows above bind.}
  \label{tab:fidelity}
  \footnotesize
  \begin{tabular}{@{}l r r r r r r r r@{}}
    \toprule
    & & \multicolumn{2}{c}{Faithful and bound} & \multicolumn{4}{c}{Decided} & Decided, no \\
    \cmidrule(lr){3-4}\cmidrule(lr){5-8}
    Configuration & Faithful & n & share & n & holds & violated & share & restatements \\
    \midrule
    Full \aratest               & 55 & 54 & \qty{22.2}{\percent} & 40 & 32 & 8 & \qty{16.5}{\percent} & 35 \\
    $-$ Source-preserving readers & 53 & 52 & \qty{21.4}{\percent} & 38 & 30 & 8 & \qty{15.6}{\percent} & 34 \\
    $-$ Source contracts$^{\dagger}$ & 44 & 43 & \qty{17.7}{\percent} & 31 & 24 & 7 & \qty{12.8}{\percent} & 31 \\
    $-$ Numeric skeletons$^{\dagger}$ & 18 & 17 & \qty{7.0}{\percent}  & 13 & 11 & 2 & \qty{5.3}{\percent}  & 13 \\
    \bottomrule
\end{tabular}

\end{table}

Full \aratest bound 54 claims (\qty{22.2}{\percent}) and decided 40
(\qty{16.5}{\percent}): 32 held and eight were violated
(\autoref{tab:fidelity}). We widened no tolerance to obtain a decision, so
14 bound claims remained \vinconclusive. Most decisions test a relation over
evidence rather than match a restated summary. Only five depend on reported
summaries alone (Andes C02 and C03, Venn C01, Triton C05, and chat-RL C05),
and excluding them leaves 35 decided claims (\qty{14.4}{\percent}). Replay
of all 187 reported-property records recomputed every decision from its
saved observations without mismatch or drift.

The relation catalog contributes most of this coverage. We removed the
components cumulatively (\autoref{tab:fidelity}). Without source-preserving readers,
\aratest bound two fewer claims. Removing source contracts then cost
nine more, two of them the RE-Bench claims that the source-format loader
binds, and removing numeric skeletons cost 26 more, which left 17 bound
claims.

With the base catalog alone, missing relations limited coverage more than
the agent did. The agent declared only 24 claims, and about half of its 219
skips (111) cited a missing relation, often because the claim asserts a
magnitude, such as a 31.3-fold cost reduction, rather than an ordering.
Numeric skeletons supply these magnitude checks. Review mattered as much as
relations did. Because review requires every substantive number and condition
of the claim, it accepted only 18 of the 24 declarations as faithful. Before
any check, it had withdrawn five of the other six because they changed their
claim, for example by replacing a rounded ratio with an inequality.

To determine whether each verdict reflects the source paper or the compiled
artifact, we traced the eight violated and 14 inconclusive claims to their
source papers (\autoref{tab:triage}). For each claim, a fresh-context AI
reviewer downloaded the cited, latest, and camera-ready versions,
transcribed the source values, and redid the deciding arithmetic. A
separate AI agent recomputed the two paper errors below. The triage changes
no RQ1 count.

\begin{wraptable}{r}{0.45\columnwidth}
  \wraptop
  \centering
  \caption{Source triage of the 22 RQ1 claims that \aratest found
    violated or left inconclusive.}
  \label{tab:triage}
  \wrapcapskip
  \footnotesize
  \setlength{\tabcolsep}{3pt}
  \begin{tabular}{@{}lrr@{}}
    \toprule
    Source of the outcome & Violated & Inconcl. \\
    \midrule
    Prose contradicts its own table & 2 & 0 \\
    Paper mismatch or overstatement & 2 & 2 \\
    Compiled artifact or declaration & 4 & 4 \\
    Released raw data: holds & 0 & 1 \\
    Undecidable at printed precision & 0 & 7 \\
    \lightmidrule
    Total & 8 & 14 \\
    \bottomrule
\end{tabular}

  \wrapbottom
\end{wraptable}

The violations expose defects on both sides of the compilation step. Four
trace to the source papers: two numerical errors, for which we found no
prior report, and two smaller mismatches, in a table caption and in the NanoGPT
speedrun's released data. \citet{zhao2024apt}
state that \apt trains 8.4$\times$ faster than its pruning baseline on
RoBERTa and 8.2$\times$ faster on T5, but the ratios in their Table~2 are
8.66$\times$ and 9.33$\times$, and the raw times in their Table~11 give
8.66$\times$ and 8.13$\times$. The stated values match the coarser Table~2
of the first arXiv version. \citet{knappe2025semantic} state that centroid
proximity weighting gains 3.14 and 0.97 points on AQuA-RAT and SVAMP, but
their Table~1 averages to 2.44 and 0.91. The stated values match the
averages of the paper's other method, but for SVAMP only after we recompute
one printed gain from its accuracies. After our disclosure, the authors of
both papers confirmed these errors. We checked only these numbers, not the
papers' other conclusions. The other four violations come from the compiled artifact. The
\CompilerSkill skill swapped two \sema columns (\autoref{sec:eval-repeat}) and made
two hedged claims strict, and Restricted-MLM C11 states the artifact's own
analysis rather than a paper claim.

The two paper errors show why an executable property catches defects that
careful reading misses. In both, a sentence states a number that the
paper's own table contradicts, but the number appears nowhere in the table,
so a reader sees no mismatch without redoing the arithmetic. The \ara keeps
the sentence and the table apart: \Statement holds the sentence, and the evidence that \Proof
links holds the cells. The declaration turns each stated number into
arithmetic over those cells, and the checker evaluates it over the interval
that each printed value's precision allows. For \apt, the ratio lies in
$[8.660, 8.662]$, outside the band $[8.35, 8.45]$ that a stated 8.4
permits. Interval arithmetic thus rules out rounding as the cause of the
gap. Because an \ara transcription error could also cause one, the triage
checked each cell against the source paper before blaming the paper.

Half of the inconclusive verdicts reflect printed precision, not a gap in
the checker (\autoref{tab:triage}). Seven claims agree with every printed number but
state a value derived from two rounded numbers, such as a 21.8-point gain
computed as $79.1-57.3$. Each operand is known only to $\pm0.05$, so the
difference is known only to $\pm0.1$, wider than the stated $\pm0.05$
band. Only unrounded results or reruns can decide them. Of the other seven,
two hide small overstatements in their papers, four hold once we fix an
artifact misreading or read the declaration as the paper wrote it, and the
authors' released raw data decide one.

Agreement on sampled inputs also left the two author CPU helpers
\vinconclusive, although we compared them with independent references on 46
registered inputs and all 83 saved checks replayed without drift. Agreement
cannot establish universal correctness, so these checks add no decision.

The 189 unbound claims show where property testing stops. Most of them
need a method other than property testing, not another relation. The first
applicable audit label assigns 123 to another verification method, 49 to
unresolved meaning, 16 to missing or unusable evidence, and one to the
unexecuted MMD-FUSE invariant. We froze the catalog rather than add
relations for individual claims, and we keep every claim in the
denominator rather than count each gap as a missing feature.

\subsection{RQ2: Seeded-fault detection on common evidence}
\label{sec:eval-detection}

In RQ2, we ask whether the deterministic checker detects seeded faults that
LLM judges miss when every method receives the same evidence. We report fault
detection, packet verdict correctness, and false alarms on valid controls
separately for each population, because a correct verdict can cite a defect
other than the injected one.

\begin{wraptable}{r}{0.5\columnwidth}
  \wraptop
  \centering
  \caption{RQ2 results from each method's first trajectory. Witness counts
    are individual submitted witnesses. Seconds average all checking
    attempts on all 508 packets, including the four executed-code packets
    and failed attempts.}
  \label{tab:detection}
  \wrapcapskip
  \footnotesize
  \setlength{\tabcolsep}{3pt}
  \begin{tabular}{@{}lrrr@{}}
    \toprule
    Endpoint & Checker & Direct & Calc. \\
    \midrule
    \multicolumn{4}{@{}l}{\emph{Recorded evidence: 504 packets, 553 faults}} \\
    Faults detected & 553/553 & 342/553 & 319/553 \\
    Correct packet verdicts & 504/504 & 458/504 & 385/504 \\
    Valid submitted witnesses & \num{1434}/\num{1434} & 565/833 & 464/579 \\
    Complete \vholds predictions & 19/19 & 17/47 & 12/35 \\
    False alarms on controls & 0/19 & 0/19 & 0/19 \\
    Operational failures & 0 & 6 & 53 \\
    \lightmidrule
    Mean seconds per attempt & 1.66 & 15.11 & 16.81 \\
    \bottomrule
\end{tabular}

  \wrapbottom
\end{wraptable}

The same evidence did not let the LLM judges match the checker. On recorded
evidence, the checker detected all 553 faults, whereas the direct and
calculator-assisted judges detected \qty{61.8}{\percent} and
\qty{57.7}{\percent} (\autoref{tab:detection}). No method raised a false
alarm on the 19 valid controls, and wherever a judge's first trajectory
disagreed with the checker, the checker was correct. The checker was also
nine to ten times faster per attempt. On the small executed-code population,
all three methods detected all three faults and returned the correct
verdict on all four packets. These counts describe two
exposed papers and one engineering helper, so we report them
descriptively, without intervals.

\begin{figure}[t]
  \centering
  \includegraphics[width=\linewidth]{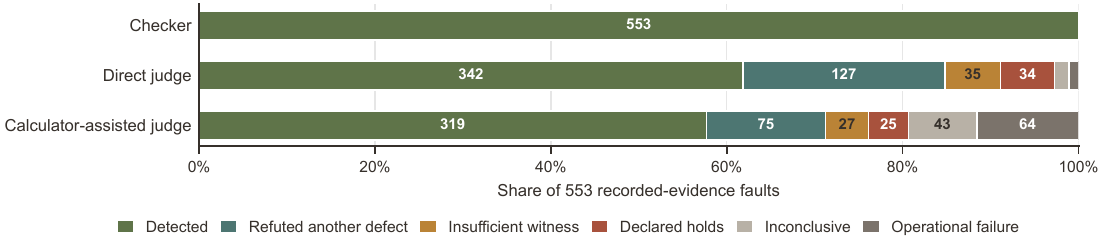}
  \caption{Primary outcome for each of the 553 recorded-evidence faults.
    ``Refuted another defect'' marks a sufficient violation that never
    identifies the injected fault.}
  \Description{One normalized stacked-bar chart over 553 faults. The checker
    detects all 553. The direct judge detects 342, refutes another defect for
    127, gives an insufficient witness for 35, declares the property holds
    for 34, returns inconclusive for 9, and fails operationally for 6. The
    calculator-assisted judge detects 319, refutes another defect for 75,
    gives an insufficient witness for 27, declares holds for 25, returns
    inconclusive for 43, and fails operationally for 64.}
  \label{fig:detection-outcomes}
\end{figure}

Finding a violation is not the same as finding the fault, and this
difference explains many judge misses (\autoref{fig:detection-outcomes}).
Most direct-judge misses (127 of 211) and a third of calculator-assisted
misses (75 of 234) correctly refuted the property through a different
comparison. One direct response cited 10.9 against 11.0, for example, while
the fault concerned 29.8 against 30.8. Both judges also declared the
property held despite a listed counterexample, as when a direct response
accepted 84.75 against 84.77, a shortfall larger than the 0.005
\presolution gate. All 26 witness errors that the scorer labeled arithmetic
were margins of the correct magnitude and opposite sign. Because our
prompt did not define the sign of a margin, we attribute them to the
response contract, not to subtraction.

Calculator access did not raise detection, but the judge rarely used the
tool's result. Only 44 of 504 first calculator trajectories called the
tool, and 37 of those stopped when the continuation exceeded the input
limit. On the 447 packets where both judges completed, the
calculator-assisted judge detected slightly more faults (318 against 307 of
484) but returned fewer correct verdicts (383 against 415), so its overall
detection gap comes from packets that only the direct judge completed.
Because the
calculator-assisted judge chose whether to call the tool, this comparison measures tool access
under matched limits, not the effect of arithmetic itself.

\begin{wraptable}{r}{0.52\columnwidth}
  \wraptop
  \centering
  \caption{RQ2 fault detection by subject (first trajectories).}
  \label{tab:breadth}
  \wrapcapskip
  \footnotesize
  \setlength{\tabcolsep}{3pt}
  \begin{tabular}{@{}lrrrrr@{}}
    \toprule
    & & & \multicolumn{3}{c}{Faults detected} \\
    \cmidrule(l){4-6}
    Subject & Packets & Faults & Checker & Direct & Calc. \\
    \midrule
    \sema & 278 & 288 & 288 & 207 & 172 \\
    \cfgartifact & 226 & 265 & 265 & 135 & 147 \\
    \lightmidrule
    Recorded evidence & 504 & 553 & 553 & 342 & 319 \\
    Triton scan helper & 4 & 3 & 3 & 3 & 3 \\
    \bottomrule
\end{tabular}

  \wrapbottom
\end{wraptable}

Unlike a property's verdict, a judge's verdict does not reproduce. Over
five trajectories, the direct judge returned one consistent verdict on 396
of 504 packets and the calculator-assisted judge on only 259, and
per-trajectory detections varied by up to 14 and 30 faults. Neither judge
dominates the other. The direct judge detected more \sema faults, and the
calculator-assisted judge more \cfgartifact faults (\autoref{tab:breadth}).
Both judges detected a larger share of held-out faults (227 and 211 of 360)
than of development-exposed faults (115 and 108 of 193). This split is over
mutations of the same two papers, not over new papers.

\begin{figure}[t]
  \centering
  \includegraphics[width=\linewidth]{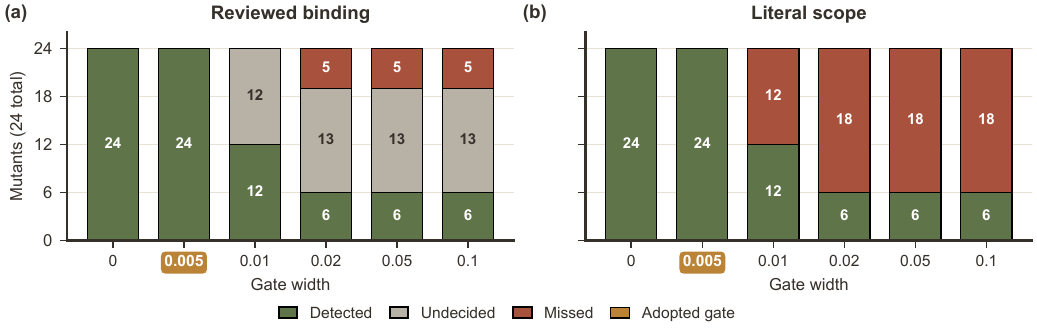}
  \caption{Gate-width sweep over 24 mutants and their unedited originals.}
  \Description{Two stacked-bar panels show that both specifications detect all
    24 mutants at gate widths zero and 0.005, 12 mutants at 0.01, and six mutants
    at wider gates. Reviewed binding routes the remaining mutants to undecided
    and missed outcomes, whereas literal scope routes all remaining mutants to
    missed outcomes.}
  \label{fig:gate-sweep}
\end{figure}

Like the choice of scope, the choice of tolerance determines which
contradictions count, so we fixed the \presolution gate from displayed
precision (\autoref{sec:catalog}) before generating the 553 mutants. To
measure that dependence, we checked 24 of the mutants and their
unedited originals at six gate widths, under the reviewed
specification and a literal-scope specification (\autoref{fig:gate-sweep}). Both detected
every mutant up to the \presolution width of \num{0.005}. At one displayed
unit (\num{0.01}), floating-point error split the 18 one-unit mutants and
hid 12, and from \num{0.02} only the six value swaps remained detected. No
original was flagged at any width, so the narrow gate cost no false alarm,
and a wider one only hides faults smaller than itself. These results
support reviewed declarations for this class of table claims, not automatic
interpretation of arbitrary claims.

\subsubsection{Replay and input drift}
\label{sec:eval-repeat}

Replay recomputes every retained RQ2 verdict offline, without a provider
call or author execution. In a clean process, it recomputed all 508 packets
and \num{3382} saved property records with no source drift.

Replay also exposes input drift. A literal-scope check on the unedited \sema
originals, outside the seeded study, flagged a claim that its compiled table did
not support. We traced the finding to the \CompilerSkill skill~\cite{liu2026ara},
which had swapped the average and final accuracy headers without failing
its schema validation. The witness replayed against the original compiled
table but diverged once we corrected the headers, and the source-bound
comparison then held on 42 of 42 comparisons. A diverging witness thus shows
when a verdict no longer applies to a corrected artifact.

\subsection{RQ3: Scope-preserving experimental workflow}
\label{sec:eval-execution}

In RQ3, we ask whether auto-research generators can draw fresh cases from
author code without leaving the claim's scope. Each test runs one generator family
on the unmodified author code and checks the claim's property on the runs it
returns, with the three verdicts of \autoref{sec:contract}. Each test also
registers control runs
whose outcome is known in advance, such as a label shuffle that must lower
the rejection rate. A passing control shows that the harness reproduces known
behavior and can detect a change, so a \vholds verdict does not come from a
run that ignored its intervention.

\begin{table}[t]
  \caption{Continuous testing on artifacts: measurements, registered rules, and controls.}
  \label{tab:execution-families}
  \footnotesize
  \begin{tabular}{@{}ll
    >{\raggedright\arraybackslash}p{0.275\linewidth}r
    >{\raggedright\arraybackslash}p{0.225\linewidth}l@{}}
    \toprule
    Artifact & Generator & Measurement and registered rule & Runs & Required controls & Verdict \\
    \midrule
    \multirow{10}{*}{\mmdfuse} & seeds & Mean rejection $0.2075$ in $[0.13,0.23]$ (one seed reached \num{0.235}) & 3 &
      Label shuffle: rejection $\leq0.10$ and drop $\geq0.05$ (observed rejection \num{0.075}) & \vholds \\
    & row order & Mean rejection \num{0.185} in $[0.13,0.23]$ & 3 &
      Same label-shuffle gates (observed rejection \num{0.075}) & \vholds \\
    & null level & Mean rejection $0.0525$ in $[0.020,0.085]$ & 3 &
      Unpermuted mixture rejects at rate $\geq0.13$ (observed 36/200) & \vholds \\
    & temperature & Paper-stated minus released: $-0.0415$, exact \qty{95}{\percent} interval $[-0.0528,-0.0301]$. Required interval inside $(-0.05,0.05)$ & 46 &
      Record-exact 36/200 reproduction. Both null counts in $[8,34]$/400 (observed 19 and 20) & \vinconclusive \\
    \midrule
    \multirow{6}{*}{\simformer} & seeds & C2ST \num{0.603}, \num{0.592} (registered mean band $[0.581,0.641]$) & 3 &
      Tenfold smaller training set raises C2ST by $\geq0.05$ (observed C2ST \num{0.904}) & \vholds \\
    & pool order & Paired C2ST changes $+0.013$ and $+0.018$. The descriptive rule gives no population verdict & 6 &
      Identity in reproduction band. Broken pairing raises C2ST by $\geq0.05$ in each block (observed about \num{0.99}) & \vinconclusive \\
    \bottomrule
\end{tabular}

\end{table}

Four of the six tests returned \vholds, and none returned a violation
(\autoref{tab:execution-families}): the \mmdfuse seed, row-order, and
null-level tests and the \simformer seed test. The property, not a single run,
decided each verdict. One \mmdfuse seed rejected at \num{0.235}, above the
band $[0.13,0.23]$, but the claim concerns the mean rate over
seeds, which stayed inside the band. The property therefore treated this
seed as one of three measurements to average, where a per-seed check would
have reported a violation.

Fresh execution also exposed two behaviors that the reported evidence could
not show, and the registered rules kept each from becoming an unsupported
violation. In \simformer, reordering the pool with pairs and partitions
intact raised C2ST by up to $+0.018$. Its registered rule describes such
changes without a bound, so its verdict is \vinconclusive by design. In
\mmdfuse, the paper-stated temperature lowered the rejection rate by
\num{0.0415} compared with the released one over \num{4000} matched pairs.
The interval of the difference excludes zero, so the two temperatures
differ, but it crosses the registered equivalence margin of $\pm0.05$, so
the rule cannot decide whether that difference violates the property.

A paired reported/fresh property on the released MMD-FUSE and MMD-Median
implementations also returned \vholds, which shows that one rule applies to
both evidence sources. Because the operands agree, the result does not show
that the rule separates them.

\begin{table}[t]
  \caption{Continuous testing campaign: the complete batches
    that the service adopted. luna is \texttt{gpt-6-luna} and GLM
    is \texttt{glm-5.3-flash}. $^{\dagger}$The baseline arm required a small
    author-code change.}
  \label{tab:continuous-campaign}
  \footnotesize
  \begin{tabular}{@{}l
    >{\raggedright\arraybackslash}p{0.50\linewidth}ll@{}}
    \toprule
    Artifact & Property and observation & Model & Verdict \\
    \midrule
    \multirow{4}{*}{\mmdfuse C12}
    & Seeds: mean rejection \num{0.2075} in $[0.13,0.23]$ (registered seed 43 reached \num{0.235}); label shuffle \num{0.075} & luna & \vholds \\
    & Seeds: the same registered seed and values & GLM & \vholds \\
    & Row order: new order, rejection \num{0.185} & luna & \vholds \\
    & Row order: registered order, rejection \num{0.19} & GLM & \vinconclusive, then \vholds \\
    \midrule
    \multirow{2}{*}{\pinn C06}
    & Convection: NNCG lowers $\log_{10}$ test L2RE below GD's by $\geq0.005$ (seeds: \num{0.04}, \num{0.34}, \num{0.20}) & luna & \vholds \\
    & Convection loss, reaction loss, and reaction L2RE: no cases returned & luna & \vinconclusive \\
    \midrule
    \multirow{2}{*}{\smm C05}
    & $\delta\odot f_\text{mask}$ beats $f_\text{mask}$ on GTSRB by $+22.5$, $+23.3$, and $+14.5$ points$^{\dagger}$ & luna & \vholds \\
    & Three attribute channels beat one: no cases returned & luna & \vinconclusive \\
    \bottomrule
\end{tabular}

\end{table}

The continuous testing campaign also returned no violation, and
operations, not the properties, limited how much it tested
(\autoref{tab:continuous-campaign}). Only seven of the 60 runs that the
service scheduled returned complete reports, and those yielded 11 checks: six
held and five were \vinconclusive. The others failed in the agent
or the container runtime, were canceled, or did not finish within the
campaign window on a server shared with other users' jobs. The complete batches also drew few new cases. On \mmdfuse, both models
reproduced the seed and row-order cells of the reviewed tests, but in three
of the four holding batches the agent reused the registered seed or row
order from the artifact's evidence. On \pinn and \smm, the holding checks
match the claims on three seeds each, although the \smm baseline arm
required a small author-code change and so does not test unmodified code.
No adopted batch covers the third \smm subclaim, C03.

Fresh execution again exposed behavior that the reported evidence could
not show. In five runs under both models, the released NNCG (NysNewton-CG)
optimizer of
\pinn crashed inside its Nystr\"om preconditioner when a Cholesky
factorization or an SVD failed. On the released reaction seed, which the
declaration requires, NNCG crashed at all five damping values in all four
attempts, so the two reaction properties cannot reach a verdict on our
software stack (the authors pin CUDA 11.7, and we used CUDA 11.8). We kept
the seed rather than drop it to obtain a verdict.

Within these limits, our RQ3 results show that auto-research generators can
run experiments on unmodified author code, both under AI-reviewed designs and
in a campaign that agents ran end to end.
Fresh execution exposed behaviors that the reported evidence
could not show, such as the \mmdfuse paper stating a temperature that its
released code does not use. Because no registered rule decided these
behaviors, we report them as \vinconclusive, not as
violations. These results do not show that agent-reviewed declarations would
survive independent human review (\autoref{sec:eval-threats}).

\subsection{Threats to validity}
\label{sec:eval-threats}

\aratest targets agent-native research, in which agents write,
maintain, and review the \ara. We evaluated \aratest under the same
condition. Fresh-context AI reviewers, not human experts, checked the RQ1
and RQ2 declarations, the six RQ3 designs, the campaign declarations, and
the paired case's declaration. AI agents also recomputed the two paper
errors and performed the \sema source check. The source authors'
confirmation of the two paper errors is the only human review of any
finding. No human checked the
campaign's runs and verdicts or the seeded mutants. Our results therefore
show how \aratest performs in an agent-maintained workflow. They do not
show that its declarations, designs, or verdicts would survive independent
human review, and we report no agreement measure between reviewers.

RQ1 coverage and most RQ2 detection concern reported evidence. Because we
designed the catalog and readers on the 243 claims, RQ1 coverage bounds
what the current catalog reaches on this corpus and may overstate its reach
on new papers. Five decisions rely only on reported summaries.
The \num{553}-of-\num{553} result in RQ2 assumes faithful declarations; it
does not measure how reliably a model constructs them. An author could
tailor a declaration to omit hard cases. Because \aratest records
the declaration's scope, evidence bindings, and provenance, a reviewer can
inspect that choice, but nothing guarantees that the reviewer notices the
omission. A passing verdict therefore does not certify a research claim.
The judges used one
mutable model alias, \texttt{gpt-6-luna}, under input and output limits
that we matched across arms but that stopped 37 first calculator
continuations; a larger allowance might change the calculator result.
Automated validators produced the reference labels and shared the
checker's evidence parser. The gate-width sweep
and the shift magnitudes vary only two coordinates of one setting, not wrong
denominators, transposed rows, or cross-table contradictions.
The six RQ3 tests used test-specific rules and produced
no falsification. They cannot establish unaided family design, the validity
of agent-run experiments without review, or real-code generative-catalog
reuse. We changed \dissclaimer's service and its agent's
instructions seven times during the campaign, so its batches did not all
run on one version. Most
of its \mmdfuse batches reused registered cases. Three artifacts from one
screening list cannot estimate how often published claims fail. The
campaign's bands and margins are study choices, not source-paper assertions.

The seeded population comes from only two artifacts
(\autoref{tab:breadth}). Four mutation operators exercise comparative
and consistency claims; \cfgartifact has a single scope column. All value
swaps and single-coordinate counterexamples were development-exposed.
Zero false alarms over 19 valid controls give weak evidence about
specificity, and because 485 of 504 recorded packets are violations, packet
accuracy alone says little. The execution population is one engineering
helper with three altered implementations on sampled inputs; it does not show
that the checker detects faults in an unmodified scientific program. The \CompilerSkill skill is stochastic, and our unseeded \sema finding
rejects the compiled relation, not the source claim.
Finally, replaying retained evidence checks a recorded computation, not a fresh
scientific execution.

\section{Discussion}
\label{sec:discussion}

A fixed declaration gives every later test the same scientific question.
Because the relation, comparison cohort, scope, and numerical policy are
explicit, users can vary admitted inputs and trace each outcome through its
witness to the evidence and interpretation behind it. A failed relation
therefore points to its binding rather than counting against the source
claim. In the agent-native workflow that \aratest targets, an agent can
carry out this testing end to end: it declares the claim, admits interventions
within the declared scope, and inspects the witness of each failure.

Many of the 189 claims that \aratest leaves unbound in RQ1 fall outside
property-based testing, and we do not expect it to verify them. General theorem proving,
static verification of all program effects, and resolution of ambiguous
scientific meaning require methods beyond property-based testing.
An executable consequence can still support a sound refutation when a
valid case violates it, even if other conditions remain untested.
A pass on that consequence cannot verify the complete assertion.
Other unbound claims, such as correlations and growth rates, fall outside only
the current catalog. We built \aratest as a composable
package for these cases. A downstream user implements a skeleton for the new
property form and registers it under a new \Kind, while the contract,
generators, and witness format stay unchanged. Until that implementation
exists, the router marks the kind \emph{pending-skeleton} and leaves the
claim untested rather than deciding it with another relation.

Generative experiments on author code exposed behaviors that the
reported evidence could not show. In \mmdfuse, the temperature that the
paper states lowered the rejection rate by \num{0.0415} compared with the
released temperature, so the paper and its code disagree on a setting that
shifts the rejection rate. The released NNCG optimizer of \pinn crashed on the seed
that the declaration requires, and reordering the \simformer pool raised
C2ST by up to $+0.018$. We report all three as \vinconclusive because no
registered rule decides whether the behavior violates the claim. That
label reflects a choice of strictness. \aratest reports a violation only
when a registered rule decides it, and leaves an exposed behavior for a
reviewer to weigh. To count such behaviors as claim defects, a
follow-up study would register the deciding rule before execution and run
unmodified controls beside each test.

We built \aratest to run while research happens, with the research agent
writing each declaration together with its claim. Our evaluation instead
added declarations to finished artifacts, so this use requires extending the
\CompilerSkill skill to emit \Kind and \Domain with each claim under an
\aisuggested tag. Agents that run experiments end to
end~\citep{lu2026aiscientist,kon2025curie} need checks from outside the
model, because LLMs struggle to correct their own reasoning without external
feedback~\citep{huang2024selfcorrect} and improve when they
receive it~\citep{shinn2023reflexion}. \aratest gives that feedback at two
speeds. Integration testing reads only stored evidence and averaged
\qty{1.66}{\second} per attempt in RQ2 (\autoref{tab:detection}), so an agent
can run it on every commit, as continuous integration gates code
changes~\citep{hilton2016ci}. Continuous testing with \dissclaimer runs fresh
experiments offline, much as continuous analysis reruns computational
workflows~\citep{beaulieujones2017continuous}, and returns failing cases as
feedback rather than as a gate. A follow-up study would measure whether this
feedback changes what the agent does next.

Beyond a single agent, \aratest could serve reviewers and downstream
artifacts. At least one venue now asks for \ara{}s alongside papers. The
\emph{Decision Sciences} special issue on decisions in the agentic era
requires one with each submission~\citep{decisci2026agentic}, so reviewers
could run \aratest on submitted claims. When one artifact relies on
another's claim, rerunning the upstream tests as the upstream evidence
changes would show which premises of the downstream work no longer hold.

\section{Related work}
\label{sec:related}

\paragraph{Specification languages and property-based testing}
\aratest's declarations play the role of program contracts and
behavioral specifications. Design by Contract attaches pre- and
postconditions to routines~\cite{meyer1992contract}, higher-order contracts
assign blame when a component breaks one~\cite{findler2002contracts}, and
JML states such obligations for Java interfaces~\cite{leavens2006jml}. A
claim contract instead specifies a relation over measurements, so it must
also fix the comparison cohort, the statistical unit, and a numeric policy
that can leave a case undecided. Probabilistic
assertions~\cite{sampson2014passert} and
\texttt{Uncertain<T>}~\cite{bornholt2014uncertain} bring uncertainty into
program specifications and types, but our numeric policy decides only when
displayed precision or a declared interval settles a comparison.
QuickCheck~\cite{claessen2000quickcheck} and Luck~\cite{lampropoulos2017luck}
type generators by value, whereas \aratest types them by
statistical unit (\autoref{sec:family-composition}).

\paragraph{Execution-backed scientific review}
Execution-backed and learned review systems address a broader semantic
problem than checking a fixed relation. FactReview resolves repositories, builds environments, and runs
bounded tasks, then uses an LLM to align outputs with reported
claims~\cite{factreview2026}. MechEvalAgent assesses coherence,
reproducibility, and generalizability through checklist-driven agents,
aggregating three runs per artifact to stabilize its
assessments~\cite{bai2026story}. \texttt{ExecutableClaims} extracts claims
and evidence and emits pytest capsules for numeric
agreement~\cite{executableclaims}. SciFact studies entailment between
scientific claims and evidence abstracts~\cite{wadden2020scifact}, while
SCITAB benchmarks reasoning over scientific tables~\cite{scitab2023}.
These systems help locate and interpret evidence. Once a declaration binds
the relation, \aratest judges new observations by a fixed test rather
than by rereading the prose. Its witness supports inspection, but a reviewer
must still check that the binding represents the source claim.

\paragraph{Declaration and evidence admission}
The declaration gate follows LLM-Modulo's proposal-and-check
design~\cite{llmmodulo2024}. Constrained decoding can enforce a structured
shape~\cite{outlines2023}, and deterministic binding checks can require
references to real evidence. Neither check establishes semantic fidelity to
the source claim. The same gap applies to agent-run experiments.
The experiment-generator agent can decline a substitution that the claim
fixes, but it cannot show that a run faithfully tests the source claim, so
we treat failing runs as feedback, not refutation.

\paragraph{Autonomous research systems}
Autonomous research systems generate hypotheses, experiments, and
interpretations~\cite{ghareeb2026robin,gottweis2026coscientist,aygun2026era}.
A survey of these systems identifies externally validated oracles as a
remaining gap~\cite{ards2026gap}, and fuzzing-inspired research motivates
searching beyond the evidence initially
reported~\cite{he2026fuzzresearch}. \aratest gives such search a fixed
target. Agents supply evidence, the registered claim contract fixes the
verdict rule, and failing agent runs return as feedback for revising the
artifact. \ara provides the host representation and
PaperBench the source corpus~\cite{liu2026ara,paperbench2025}.

\section{Conclusion}
\label{sec:conclusion}

Agents now produce research claims faster than researchers can check
them. Each claim implies a property but states it only in prose, so no test
can decide which evidence would falsify the claim. We make claims testable by
approximating their semantics with property skeletons. \aratest binds
each claim to a skeleton through a typed declaration and checks the bound
property as an executable test whose scope and decision rule stay fixed
across reported and fresh evidence.
On reported evidence, \aratest detected more seeded faults than LLM
judges in about a tenth of the time and found two published papers whose
prose contradicts their own tables, errors that their authors confirmed. On fresh runs of author code, the same
properties exposed behaviors that reported evidence could not show,
such as released code that uses a different temperature than its paper
states. Because the property stays fixed while the evidence changes,
researchers and agents can revisit a claim without silently changing what
they test. Most claims that \aratest leaves unbound need a verification
method other than property testing, and credible fault discovery will
require preregistered studies with independent review of declarations.

\section*{Ethics and privacy statement}

We analyze only published papers and their publicly released artifacts,
which contain no private personal data. We name each paper so that readers
can check our findings against it. Unlike a new software vulnerability, an
error in a published paper exposes no system to attack, so we did not
withhold our findings until the authors could respond. A finding concerns a
stated claim and its evidence, not the conduct of the source authors, and we
do not infer misconduct from a failed check. We disclosed our findings
to the authors of the papers, and the authors of both papers with
numerical errors (\autoref{sec:eval-fidelity}) confirmed them.

\printbibliography

\appendix
\renewcommand{\sectionautorefname}{Appendix}
\section{Generator interface}
\label{sec:generator-interface}

We list the generator interface of our Python library in
\autoref{fig:generator-classes}. We describe each class in
\autoref{sec:claim-scoped-generation} and the combinators in
\autoref{sec:family-composition}.

\begin{figure}[t]
  \centering
  \input{listings/generator-classes}
  \caption{Generator interface of the Python library. A
    \texttt{Generator} returns a batch of cases of type \texttt{C} with its
    coverage. \texttt{AgentGenerator} draws cases by dispatching the
    experiment-generator agent. Each experiment family type is registered
    data: an instruction, a parameters model, and a case type.
    \texttt{one\_of} and
    \texttt{frequency} combine generators that share a case type
    \texttt{C}.}
  \label{fig:generator-classes}
\end{figure}

\section{LLM judge prompt}
\label{sec:judge-prompt}

We give both RQ2 judges the same system prompt and reproduce it verbatim in
\autoref{fig:judge-prompt} and \autoref{fig:judge-schema}. The prompt states
the decision rules and fixes the response schema, but it names no claim.
We sent it as one line; here we break it at sentence boundaries and lay out
the schema one property per line. The schema leaves the sign of
\texttt{margin} undefined, which explains the opposite-sign witness errors
in \autoref{sec:eval-detection}.

A single user message follows the prompt and carries the evidence packet as
compact JSON. The packet holds a plain-language description of the property,
the observation, comparison, and clause rows that the prompt defines, and the
coverage and completeness limits. Recorded packets also carry the raw table
and claim text. The deterministic checker receives the same packet. Fault
labels, mutation locations, and reference verdicts never appear in it.

The calculator-assisted judge also receives one tool, \texttt{calculator},
described as ``Bounded exact arithmetic on supplied decimal operands; no
external access.'' A call names one of ten operations (\texttt{add},
\texttt{subtract}, \texttt{multiply}, \texttt{divide}, \texttt{sum},
\texttt{mean}, \texttt{min}, \texttt{max}, \texttt{compare}, and
\texttt{count\_ge}) and supplies up to 256 decimal operands of at most 64
characters, a threshold for \texttt{count\_ge}, and an optional summary of
at most 512 characters. The tool computes with exact rationals and returns
the exact result with a decimal rendering of up to 36 significant digits.
The direct judge receives no tool.

\begin{figure}[t]
  \centering
  \input{listings/judge-prompt}
  \caption{Decision rules in the system prompt shared by the direct and
    calculator-assisted judges.}
  \Description{The verbatim judge prompt, one sentence per line. It restricts
    the judge to the supplied property and observations, defines the
    observation, comparison, and clause rows, sets the displayed-precision
    gate, explains when a clause refutes the conjunction and when a sampled
    property stays inconclusive, and states what a violation or holding
    answer must cite.}
  \label{fig:judge-prompt}
\end{figure}

\begin{figure}[t]
  \centering
  \input{listings/judge-schema}
  \caption{Response schema that ends the system prompt in
    \autoref{fig:judge-prompt}.}
  \Description{A JSON schema for one prediction with a verdict of holds,
    violated, inconclusive, or no verdict, a list of witnesses, the checked
    comparison identifiers, and an explanation. Each witness names a
    comparison, its subject and reference values, and an optional margin.}
  \label{fig:judge-schema}
\end{figure}

\end{document}